%% file: camera-ready.tex
\documentclass{article}

\usepackage{microtype}
\usepackage{graphicx}
\usepackage{booktabs} 

\usepackage{hyperref}

\usepackage{multirow}
\usepackage{caption, subcaption}
\usepackage{arydshln}  
\usepackage{xcolor}
\usepackage{wrapfig}
\usepackage{stmaryrd}
\usepackage{microtype}
\definecolor{red}{rgb}{0.95,0.4,0.4}
\definecolor{darkgreen}{rgb}{0.3, 0.75, 0.3}

\usepackage[accepted]{icml2024}

\usepackage{amsmath}
\usepackage{amssymb}
\usepackage{mathtools}
\usepackage{amsthm}

\DeclareMathOperator*{\argmin}{arg\,min}
\usepackage[capitalize,noabbrev]{cleveref}

\theoremstyle{plain}

\theoremstyle{definition}

\theoremstyle{remark}

\usepackage[textsize=tiny]{todonotes}

\icmltitlerunning{Revisiting the Power of Prompt for Visual Tuning}

\begin{document}

\twocolumn[
\icmltitle{Revisiting the Power of Prompt for Visual Tuning}


\icmlsetsymbol{mentor}{$\dagger$}

\begin{icmlauthorlist}
\icmlauthor{Yuzhu Wang}{zjlab}
\icmlauthor{Lechao Cheng}{hfut,mentor}
\icmlauthor{Chaowei Fang}{xidian}
\icmlauthor{Dingwen Zhang}{nwpu}
\icmlauthor{Manni Duan}{zjlab}
\icmlauthor{Meng Wang}{hfut}
\end{icmlauthorlist}



\icmlaffiliation{zjlab}{Zhejiang Lab}
\icmlaffiliation{hfut}{School of Computer Science and Information Engineering, Hefei University of Technology}
\icmlaffiliation{xidian}{School of Artificial Intelligence, Xidian University}
\icmlaffiliation{nwpu}{School of Automation, Northwestern Polytechnical University}

\icmlcorrespondingauthor{Lechao Cheng}{chenglc@hfut.edu.cn}
\icmlkeywords{Visual Prompt Tuning, Token Prototypes, Ininialization}

\vskip 0.3in
]



\printAffiliationsAndNotice{}  

\begin{abstract}
Visual prompt tuning (VPT) is a promising solution incorporating learnable prompt tokens to customize pre-trained models for downstream tasks. However, VPT and its variants often encounter challenges like prompt initialization, prompt length, and subpar performance in self-supervised pretraining, hindering successful contextual adaptation. This study commences by exploring the correlation evolvement between prompts and patch tokens during proficient training. Inspired by the observation that the prompt tokens tend to share high mutual information with patch tokens, we propose initializing prompts with downstream token prototypes. The strategic initialization, a stand-in for the previous initialization, substantially improves performance. To refine further, we optimize token construction with a streamlined pipeline that maintains excellent performance with almost no increase in computational expenses compared to VPT. Exhaustive experiments show our proposed approach outperforms existing methods by a remarkable margin. For instance, after MAE pre-training, our method improves accuracy by up to 10\%$\sim$30\% compared to VPT, and outperforms Full fine-tuning 19 out of 24 cases while using less than 0.4\% of learnable parameters.
Besides, the experimental results demonstrate the proposed SPT is robust to prompt lengths and scales well with model capacity and training data size. We finally provide an insightful exploration into the amount of target data facilitating the adaptation of pre-trained models to downstream tasks. The code is available at \href{https://github.com/WangYZ1608/Self-Prompt-Tuning}{https://github.com/WangYZ1608/Self-Prompt-Tuning}.

\end{abstract}

\section{Introduction}

The prevailing practice targeting downstream-focused computer vision applications adheres to the paradigm of pre-training followed by fine-tuning. This involves initially obtaining/training a versatile, task-agnostic backbone using supervised or self-supervised instructions. The structure of this foundation model is subsequently updated and tailored to the downstream tasks. While full parameter fine-tuning can yield impressive performance (always marked as the performance ceiling for downstream task tuning), it suffers from heavy computational burden that necessitates updating all model parameters and maintaining a distinct copy for each task when serving online. These issues become exacerbated particularly for modern model architectures with ever-expanding capacity, such as ViT-H (632M paramters~\cite{dosovitskiy2020image}) and ViT-22B~\cite{dehghani2023scaling}, rendering it often impractical in real scenarios. As an alternative to trade off computational cost and performance, parameter-efficient fine-tuning typically brings part of tunable modules~\cite{zhai2019large, houlsby2019parameter, lian2022scaling, jia2022visual} that usually have fewer parameters for task adaptation. Among these approaches, visual prompt tuning (VPT~\cite{jia2022visual}) is gaining momentum as a promising solution that learns task-specific tokens while keeping the pretrained foundation model frozen during the tuning phase.
\input{figures/freamwork}
Recent studies investigating VPT~\cite{jia2022visual} and its variants~\cite{yoo2023improving, cheng2023e2vpt, tu2023visual} unveil consistent challenges posed by the construction of prompt tokens\footnote{In this paper, we refer to the inserted learnable parameters as \textit{prompt tokens}, and the input images and the intermediate features as \textit{patch tokens}.}, shedding light on the intricacies involved. We summarize the main challenges outlined below.

\hspace{1.2em} \textbf{\textit{Prompt initialization}}: Existing prompt-based methods, such as VPT~\cite{jia2022visual}, GateVPT~\cite{yoo2023improving}, employ the strategy of initializing prompts randomly (e.g., uniform or normal) and then update prompts during tuning, akin to optimizing the parameters of conventional neural networks. Nevertheless, the distinct initialization techniques for inserted prompt tokens significantly impact accuracy, as shown in the original paper of VPT and its variants.

\hspace{1.2em} \textbf{\textit{Prompt length}}: The only extra hyperparameter that requires tweaking, in comparison to full fine-tuning, is the number of inserted prompt tokens. While ablation studies of existing approaches showcase that VPT and its variants are usually sensitive to the number of inserted prompts.

\hspace{1.2em} \textbf{\textit{Subpar performance with self-supervised pretraining}}: Recent works~\cite{yoo2023improving} have proven that the performance of VPT and its variants under self-supervised pretraining is significantly worse than its performance under supervised pretraining, hindering its application in various scenarios with massive unlabeled data.

To investigate these challenges, we employ a straightforward approach by conducting simple exploratory experiments to observe the distributional relationship, Masked Autoencoder (MAE~\cite{he2022masked}) pre-trained ViT-B as the backbone. As illustrated in Fig.~\ref{fig:nmi}, we compute the normalized mutual information between prompt tokens and patch tokens in each layer.
Clearly, 
as the training proficiency progresses, the prompt tokens for downstream contextualization gradually converge towards the distribution of patch tokens, manifesting specifically as an increase in mutual information.

In light of the observation, we hypothesize that if the prompt tokens share sufficient information with patch tokens during the initial stage, the visual prompt tuning converges more rapidly and achieves superior results on the target task. To this end, we leverage the downstream inferred token prototypes as the initial prompts. Specifically, we first fed the training images into the pre-trained backbone for the target task to generate a collection of inferred tokens and cluster them together based on the corresponding prompt length. Subsequently, we initialize the task-specific learnable prompt tokens with those prototypes (Fig.~\ref{fig:freamwork}). The results on the benchmark datasets demonstrate dramatic improvements with our proposed approach (Tab.~\ref{tab:sota_ssl}). However, the time cost associated with the clustering of tokens is extraordinarily substantial, bordering on intolerable. To further optimize the procedure, we propose to sample the inferred tokens as the initial ones based on several sampling strategies, e.g. mean pooling, max pooling, and random sampling, achieving compelling performance with nearly no extra cost (Tab.~\ref{tab:time-cost}).

The proposed SPT establishes a solid training start and is robust to prompt length, striking a trade-off between performance and computational cost. It positions itself as a robust complement to VPT for the adaptation of downstream tasks. We summarize the contributions as follows:

\begin{itemize}
    \item We propose a simple yet effective approach termed \textbf{S}elf-\textbf{P}rompt \textbf{T}uning (SPT) that leverages the downstream inferred token prototypes as the initial prompts inspired by high mutual information. Additionally, we accelerate the clustering procedure by randomly sampling inferred tokens, which results in significant improvements with minimal additional computational cost compared to naive VPT. The proposed SPT is robust to prompt length and scales well with model capacity and training data size.
    \item For the first time, we showcase that the proposed SPT can achieve highly accurate results and can even compete with full fine-tuning. For instance, after MAE pre-training, our method improves average accuracy by up to 10\% $\sim$ 30\% relative to VPT, and outperforms full fine-tuning in 19 out of 24 cases while using less than 0.4\% of learnable parameters. 
\end{itemize}

\section{Related Work}
\textbf{Pre-training and Full Fine-tuning.} 
Pioneered by the seminal work of ImageNet~\cite{deng2009imagenet}, the methodology for image classification and various other computer vision tasks typically involves a pre-training followed by fine-tuning approach. In this paradigm, a general-purpose, task-agnostic backbone undergoes pre-training~\cite{krizhevsky2012imagenet, dosovitskiy2020image} through supervised or self-supervised~\cite{he2020momentum} methods. Subsequently, the structure of this pre-trained backbone is modified and adapted to specific downstream tasks. 
Recent studies show that self-supervised pre-training~\cite{he2020momentum, grill2020bootstrap, chen2021mocov3, chen2020simple, he2022masked} in downstream tasks outperforms supervised~\cite{goyal2017accurate, he2019rethinking} or weakly supervised~\cite{kolesnikov2020big} pre-training in full fine-tuning and shows promising scaling behavior. However, the results are always suboptimal~\cite{jia2022visual, yoo2023improving} when conducting visual prompt tuning on self-supervised pretrained models. The proposed method in this work improves the accuracy of VPT, especially for self-supervised pre-trained models, thereby effectively mitigating the limitation of VPT.


\textbf{Parameter-Efficient Fine-Tuning.}
\textit{Full fine-tuning} is widely used to tailor the pre-trained foundation model to the downstream task. It is a straightforward approach that can often deliver impressive results, but it comes at the expense of updating all parameters and necessitates storing a distinct fine-tuned model for each task. As an alternative, \textit{linear probing} focuses solely on training and storing new task heads, keeping the backbone frozen. However, its performance tends to lag behind that of full parameter fine-tuning. To strike a better balance between computational cost (i.e., training time and memory usage) and performance, researchers are exploring parameter-efficient fine-tuning (PEFT) algorithms. 
For example, SpotTune~\cite{guo2019spottune} investigates which layers require fine-tuning, while Bitfit~\cite{zaken2021bitfit} updates only the bias term. SSF~\cite{lian2022scaling} proposes scaling and shifting deep features, and some works advocate inserting adapters~\cite{houlsby2019parameter, chen2022adaptformer} into the network. Different from these methods of tuning backbones, VPT~\cite{jia2022visual} draws inspiration from prompt tuning~\cite{li2021prefixtuning, lester2021power, liu2022ptuning} in NLP to introduce a set of learnable prompts tokens into the input space and optimizing them while keeping the backbone frozen, leading to significant computational saving. Surprisingly, VPT achieves comparable results even on par with Full fine-tuning. However, existing studies~\cite{yoo2023improving} also demonstrate that VPT and its variants suffer from some challenges, as discussed before. This work attempts to initialize the inserted prompt tokens with high mutual information, and the results demonstrate the proposed approach can alleviate those issues. We believe that the proposed technique can be viewed as a  substantial  complement to the original VPT framework.

\section{Methods}\label{Sec:methods}

Within this part, we first revisit the standard visual prompt tuning (VPT~\cite{jia2022visual}). Next, we conduct a simple exploratory experiment to investigate the connections between prompt tokens and patch tokens as the training proficiency progresses. Upon careful inspection, we finally detail the proposed approach and optimize the process.

\subsection{Visual Prompt Tuning}\label{sec:vpt}

Over the past few years, Vision Transformers (ViT)~\cite{dosovitskiy2020image} have been established as a powerful backbone for visual recognition. ViT usually consists of a patch embedding layer, a stack of $L$ transformer blocks, and a classification head. The patch embedding layer divides the input image $\mathbf{x}$ into a set of non-overlapping patches. Each patch is then embedded into $D$-dimensional latent space coupled with the position information, as follows:
\begin{equation}
    \mathbf{E}_0 = {\rm PatchEmbed}(\mathbf{x}) + \mathbf{E}_{pos}
\end{equation}
\hspace{1.2em} The image patch embeddings, $\mathbf{E}_i = \{\mathbf{e}_i^j \in \mathbb{R}^D | j \in \mathbb{N}, 1\leq j \leq N_e \} $, together with an extra learnable class token ([CLS]), as inputs to the ($i$+1)-th Transformer block (B$_{i+1}$), where each block consists of a multi-head self-attention~\cite{vaswani2017attention} followed by a feed-forward layer with layer normalization~\cite{ba2016layer} and residual connection~\cite{he2016deep}. The whole ViT is formulated as:
\begin{equation}
    [\mathbf{x}_i, \mathbf{E}_i] = {\rm B}_i([\mathbf{x}_{i-1}, \mathbf{E}_{i-1}]),   i=1,2,\cdots,L
\end{equation}
\begin{equation}
    \mathbf{y}={\rm Head}(\mathbf{x}_{L})
\end{equation}
\hspace{1.2em} where $\mathbf{x}_i\in\mathbb{R}^D$ denotes [CLS]'s embedding at ($i$+1)-th block input.  [$\cdot$, $\cdot$] indicates stacking and concatenation on the sequence length dimension. The task head is used to map the last block's [CLS] embedding, $\mathbf{x}_L$, into a predicted class probability distribution $\mathbf{y}$.\footnote{Masked Autoencoder (MAE) pretraining does not use [CLS] token. We follow the original designs and treat global average pooled image embedding as input for the task head.}

\noindent\textbf{VPT-Shallow}
In VPT framework proposed by Jia et.al~\cite{jia2022visual}, only a set of $N_p$ learnable prompt tokens, denoted as $\mathbf{P}_{0} = \{\mathbf{p}_{0}^{k} \in \mathbb{R}^D | k \in \mathbb{N}, 1 \leq k \leq N_p\}$, is introduced in the first transformer block. The VPT-Shallow is formulated as:
\begin{equation}
    [\mathbf{x}_1, \mathbf{Z}_1, \mathbf{E}_1] = {\rm B}_1([\mathbf{x}_{0}, \mathbf{P}_{0}, \mathbf{E}_{0}])
\end{equation}
\begin{equation}
    [\mathbf{x}_i, \mathbf{Z}_i, \mathbf{E}_i] = {\rm B}_i([\mathbf{x}_{i-1}, \mathbf{Z}_{i-1}, \mathbf{E}_{i-1}]),   i=2,\cdots,L
\end{equation}

\noindent\textbf{VPT-Deep}
For a more complicated setting, we insert $N_p$ trainable prompt tokens $\mathbf{P}_{i-1} = \{\mathbf{p}_{i-1}^{k} \in \mathbb{R}^D | k \in \mathbb{N}, 1 \leq k \leq N_p\}$ into $i$-th transformer block\footnote{For convenience, we set the inserted prompt length to $N_p$ in each layer for better concatenation. If we were to devise tokens with well-crafted dynamic prompt lengths, it is worth noting that this is often non-trivial and falls beyond the scope of this study.}. The VPT-Deep is defined as:
\begin{equation}
    [\mathbf{x}_i, \_, \mathbf{E}_i] = {\rm B}_i([\mathbf{x}_{i-1}, \mathbf{P}_{i-1}, \mathbf{E}_{i-1}]),   i=1,\cdots,L
\end{equation}

\noindent\textbf{Discussion.} VPT showcases considerable potential in downstream contextualization, leveraging large-scale pre-trained models. Deviating from conventional full parameter fine-tuning, VPT efficiently addresses the significant computational cost (updating less than 1\% of parameters) and storage overhead (storing a single large foundation model and a small set of tokens coupled with a task-specific head for each task). It notably achieves competitive performance, retains the generalization capabilities of the foundation model, and alleviates overfitting within a certain amount of data. However, current advancements highlight substantial fragility when applying VPT to downstream tasks with different settings for prompt initialization and length. Also, the performance of the more frequently used self-supervised pre-trained models appears not to be as good as we expected.

\subsection{High Normalized Mutual Information Matters}
The central concern has pivoted towards optimizing the training of inserted prompt tokens. We start by conducting a simple experiment to observe how the distribution of prompt tokens changes as VPT converges. An intuitive strategy involves utilizing patch tokens as a reference. Consequently, we introduce \textbf{N}ormalized \textbf{M}utual \textbf{I}nformation (NMI~\cite{estevez2009normalized}) as a metric to quantify the relationship. Let $\pi(\textbf{p}_{i-1}, \textbf{e}_{i-1})$ be the joint distribution of prompt and patch tokens for the $i$-th transformer block. We approximate $\pi$ with sigmoid-normalized cross-attention over $\mathbf{P}_{i-1}$ and $\mathbf{E}_{i-1}$ as:

\begin{equation}\label{eq:joint}
\pi(\textbf{p}^{k}_{i-1}, \textbf{e}^{j}_{i-1}) \coloneqq \frac{\sigma(\mathbf{p}^{k}_{i-1} \cdot \mathbf{e}^{j}_{i-1} )}{\sum_{j,k}\sigma(\mathbf{p}^{k}_{i-1} \cdot \mathbf{e}^{j}_{i-1})}
\end{equation}
where $\sigma$ is the sigmoid operation. We further collapse along $j$ to derive the marginal distribution of $\textbf{p}_{i-1}$ as $\pi(\textbf{p}_{i-1})=\sum_{j=1}^{N_e}\pi(\textbf{p}^{k}_{i-1}, \textbf{e}^{j}_{i-1})$. The marginal distribution of $\textbf{e}_{i-1}$ follows the same principle as $\pi(\textbf{e}_{i-1})=\sum_{k=1}^{N_p}\pi(\textbf{p}^{k}_{i-1}, \textbf{e}^{j}_{i-1})$. Thus, the normalized mutual information is defined as:
\begin{equation}\label{eq:nmi}
\begin{aligned}
    NMI(\mathbf{P}_{i-1};\mathbf{E}_{i-1}) = \frac{2*I(\mathbf{P}_{i-1};\mathbf{E}_{i-1}) }{H(\mathbf{P}_{i-1}) + H(\mathbf{E}_{i-1})}
\end{aligned}
\end{equation}
where $I(;)$ indicates the standard mutual information, and $H(.)$ is the entropy. We can easily compute them with the joint distribution $\pi(\textbf{p}^{k}_{i-1}, \textbf{e}^{j}_{i-1}) $ and marginal distributions.

To explore the degree of attention collapse for prompt tokens, we measure the normalized mutual information for four canonical datasets (CUB-200-2011~\cite{wah2011caltech}, Caltech-101~\cite{wah2011caltech}, Patch Camelyon~\cite{veeling2018rotation}, Clevr\/count~\cite{johnson2017clevr}).
As can be found in Fig.~\ref{fig:nmi}, the training procedure of visual prompt tuning (e.g., blue curves in four sub-figures) gradually leads to high normalized mutual information values. Considering the definition of NMI itself, we conclude our discovery as follows:

\noindent\textbf{Discovery.} \textit{The distribution of prompts for downstream contextualization gradually converges towards the distribution of patch tokens as the training proficiency progresses, manifesting specifically as increasing values in normalized mutual information.}

Recall that we discussed the benefits and challenges of applying visual prompt tuning in Sec.~\ref{sec:vpt}. Given the discovery above, we conjecture that the initial prompt tokens with highly shared information (high NMI values at the training start point) may benefit the tuning procedure (improved convergence speed and accuracy), fostering more stable training. The problem is then shifted to constructing prompt initialization with high normalized mutual information.

\input{figures/nmi}

\subsection{Self-Prompt Tuning based on Token Prototypes}
The intuitive approach to constructing prompts with highly shared information is to initialize the prompt tokens with inferred token prototypes on the target dataset. We denote the forward results of patch embeddings in layer $i$ as $\mathbf{E^*}_i = \{\mathbf{e}_i^j \in \mathbb{R}^D | j \in \mathbb{N}, 1\leq j \leq N_e * N \} $ for the target dataset. $N$ is the total number of the entire target dataset. Supposing $\mathbf{C}_i = \{\mathbf{c}_i^j \in \mathbb{R}^D | j \in \mathbb{N}, 1\leq j \leq N_p\}$ are prototypes for layer $i$, we cluster $\mathbf{E^*}_i$ into $N_p$ prototypes with K-means by minimizing the inertia as:
\begin{equation}
\begin{aligned}
\argmin_{\mathbb{S}_i}\sum_{j=1}^{N_p}\sum_{\mathbf{x}^j\in \mathbb{S}_i^j} ||\mathbf{x}^j - \mathbf{c}_i^j||^2
\end{aligned}
\end{equation}
\begin{equation}
    \begin{aligned}
        \mathbf{c}_i^j =\frac{1}{\| \mathbb{S}_i^j\|} \sum_{\mathbf{x}^j \in \mathbb{S}_i^j } \mathbf{x}^j
    \end{aligned}
\end{equation}
where $\mathbb{S}_i$ are cluster sets and are initially partitioned uniformly with $\mathbf{E^*}_i$. $\mathbf{x}^j$ means the patch embedding issued into the $j$-th cluster set $\mathbb{S}_i^j$.

Similar to visual prompt tuning, we construct the prompt tokens with the prototypes $\mathbf{C}_i$. We term our approach \textbf{S}elf-\textbf{P}rompt \textbf{Tuning} (\textbf{SPT}). Following the example of VPT, we define \textbf{SPT-Shallow} and \textbf{SPT-Deep} depending on the number of transformer blocks involved.

\noindent\textbf{SPT-Shallow}
We only initialize $N_p$ learnable prompt tokens with $\mathbf{C}_{0} = \{\mathbf{c}_{0}^{k} \in \mathbb{R}^D | k \in \mathbb{N}, 1 \leq k \leq N_p\}$ in the first transformer block. The SPT-Shallow is formulated as:
\begin{equation}
    [\mathbf{x}_1, \mathbf{Z}_1, \mathbf{E}_1] = {\rm B}_1([\mathbf{x}_{0}, \mathbf{C}_{0}, \mathbf{E}_{0}])
\end{equation}
\begin{equation}
    [\mathbf{x}_i, \mathbf{Z}_i, \mathbf{E}_i] = {\rm B}_i([\mathbf{x}_{i-1}, \mathbf{Z}_{i-1}, \mathbf{E}_{i-1}]),   i=2,\cdots,L
\end{equation}

\noindent\textbf{SPT-Deep}
For the input of $i$-th transformer block, we initialize $N_p$ inserted prompt tokens with corresponding prototypes $\mathbf{C}_{i-1} = \{\mathbf{c}_{i-1}^{k} \in \mathbb{R}^D | k \in \mathbb{N}, 1 \leq k \leq N_p\}$. The SPT-Deep\footnote{For vision transformer, the positional order of tokens has no effect after positional encoding, i.e., $[\mathbf{x}_{i}, \mathbf{C}_{i}, \mathbf{E}_{i}]$ and $[\mathbf{C}_{i}, \mathbf{x}_{i}, \mathbf{E}_{i}]$ are mathematically equivalent.} is defined as:
\begin{equation}
    [\mathbf{x}_i, \_, \mathbf{E}_i] = {\rm B}_i([\mathbf{x}_{i-1}, \mathbf{C}_{i-1}, \mathbf{E}_{i-1}]),   i=1,\cdots,L
\end{equation}

We further compute the NMI based on eq.~\ref{eq:joint} and eq.~\ref{eq:nmi}. The red curves in Fig.~\ref{fig:nmi} illustrate the trend of NMI over epoch for SPT. SPT starts with large NMI values as expected, facilitating rapid convergence and convincing results. During adaptation training on downstream tasks (use the whole training data), only the parameters of prompts and task head are learnable, while the Transformer encoder is frozen. In addition, we found that after self-prompt initialization, even if the parameters of the prompts are frozen (which means that only the task head is learnable), SPT can still achieve satisfactory results (Tab.~\ref{tab:ablation_prompt_sample} (a)). That indicates we indeed obtain a very good starting point with SPT.
More experiments can be found in the experimental section.

\subsection{Optimization for Token Construction}
The previously introduced SPT treats token prototypes as initial prompts, which could approximately represent the overall distribution of the training data and/or intermediate tokens. However, the number of candidate tokens is usually far more than the length of the prompts. The clustering process is extremely slow (see Tab.~\ref{tab:time-cost} and Fig.~\ref{fig:ablation_componments} (a)), and even the time cost significantly exceeds the time of model fine-tuning. To alleviate this issue, we propose two more practical strategies.

\noindent\textbf{Mean (max) pooling} We perform mean or max pooling on one randomly selected batch tokens $\mathbf{E'^*}_i = \{\mathbf{e}_i^j \in \mathbb{R}^D | j \in \mathbb{N}, 1\leq j \leq N_e * B \}$ to output $N_p$ tokens. For non-overlapping sampling, the window kernel size is $\lfloor \frac{N_e * B}{N_p} \rfloor $. $\lfloor . \rfloor$ means floor operation. $B$ is the batch size.

\noindent\textbf{Random sample} For a streamlined alternative, we even randomly sample $N_p$ tokens within a random batch. For example, we randomly select one forward batch result of patch embeddings in layer $i$ as $\mathbf{E'^*}_i$. The final $N_p$ initial tokens are uniformly sampled from $N_e * B $ tokens.

\noindent\textbf{Remark} The proposed refinement procedure is more practical for downstream tasks. We undertake a brief assessment of computational cost in two aspects. 

\noindent\textbf{(I) Parameters}: Prompts are the only additional hyper-parameters that need to be tuned and stored for SPT compared to full parameter fine-tuning. We only store the learned prompts and task head for each downstream task and re-use the copy of the pretrained backbone, significantly reducing the storage cost. For instance, given a ViT-B with 86M parameters and embedding dimension $D=768$, we set the length of learnable prompts to 100 and 20 for shallow and deep variants respectively, which yield additional $N_p\times D=100\times 768=0.0768$M, and $L \times N_p \times D=12\times 20 \times 768=0.18$M, amounting to only 0.089\% and 0.21\% of ViT-B parameters, respectively.

\noindent\textbf{(II) Time Cost}: Although the token prototypes based approach achieves impressive performance, the time cost of clustering tokens is extremely high. For example, we even spend 27.3 days executing K-means when applying SPT-Deep with token prototypes on CUB-200-2011 (see Tab.~\ref{tab:time-cost}). Nevertheless, the proposed mean pooling, max pooling, and random sample strategies only need about 42.63 seconds, 64.86 seconds, and 43.56 seconds, respectively. Surprisingly, those two sampling strategies can still achieve compelling results (see Tab.~\ref{tab:ablation_prompt_sample} for more details).
\begin{table}[]
    \centering
    \caption{
    \textbf{The time cost of the process of constructing prompts}. 
    We employ SPT-Deep over four strategies (K-Means, Max pooling, Mean pooling, Random sample) on CUB-200-2011 with ViT-B as the backbone. (Note: 'd' means days, while 's' means seconds.)}
    \vspace{2mm}
    \resizebox{0.98\linewidth}{!}{%
    \begin{tabular}{lp{0.13\linewidth}p{0.13\linewidth}p{0.13\linewidth}p{0.13\linewidth}p{0.13\linewidth}}
    \toprule
    \multirow{2}{*}{Strategies} & \multicolumn{5}{c}{Proportion of the training data to construct prompts}\\
    \cmidrule{2-6}
    & 10\%  & 30\%   & 50\%   & 70\%   & 100\% \\
    \midrule
    K-means& 0.10d  & 0.30d   & 1.70d   & 6.20d   & 27.30d  \\
    Mean pooling& 4.83s & 13.20s & 21.80s & 29.19s & 42.63s\\
    Max pooling& 7.20s& 19.93s& 32.80s& 45.11s& 64.86s\\
    Random sample& 5.30s  & 14.65s & 24.13s & 32.50s & 43.56s\\
    \bottomrule
    \end{tabular}
    }
    \vspace{-5mm}
    \label{tab:time-cost}
\end{table}

\section{Experiments}

\subsection{Experiment Setup}\label{sec:setup}

\input{table/compare_ssl}
\input{table/Appendix_FGVC}
\noindent\textbf{Datasets and Backbones.}
Our experiments are carried out on two image classification benchmarks. All the hyperparameters are determined with cross-validation on \texttt{val} set.

\hspace{1.2em}\textbf{FGVC} contains 5 benchmarked Fine-Grained Visual Classification, including CUB-200-2011~\cite{wah2011caltech}, NABirds~\cite{van2015building}, Oxford Flowers\cite{nilsback2008automated}, Stanford Dogs\cite{khosla2011novel} and Stanford Cars~\cite{gebru2017fine}. Following VPT~\cite{jia2022visual}, we randomly split the training set into \texttt{train} ($90\%$) and \texttt{val} ($10\%$). 

\hspace{1.2em}\textbf{VTAB-1K} collects 19 benchmarked Visual Task Adaptation~\cite{zhai2019large}, categorized into three groups: i) \textit{Natural} contains natural images captured by standard cameras; ii) \textit{Specialized} images that are curated by specialized equipment; iii) \textit{Structured} which requires structural understanding such as 3D depth prediction. Each task of VTAB-1k contains 1000 training examples. Following~\cite{zhai2019large, jia2022visual}, we apply the 800-200 split of the train/val set. 

\hspace{1.2em}\textbf{Pretrained Backbones.} We use the plain vision transformer (ViT-Base/Large/Huge~\cite{dosovitskiy2020image}) as the pretraining backbones that involve fewer inductive biases, since the backbone may be trained effectively using large-scale data and/or self-supervision. Specifically, we initialize the backbone with supervised pre-trained on ImageNet-21K~\cite{deng2009imagenet} or self-supervised pre-trained on ImageNet-1K without labels~\cite{chen2021mocov3, he2022masked}, to ensure the flexibility of our method.

\textbf{Implementation Details.} For the FGVC datasets, we process images with a randomly resize crop
operation to 224 × 224 resolution and a random horizontal flip for data augmentation. For VTAB-1k, we directly resize images to 224 × 224 and don't adopt any other augmentations~\cite{zhai2019large, jia2022visual}.
We employ the AdamW optimizer with a mini-batch size of 32 for a total of 100 epochs (with a linear warm up for the first 10 epochs), and cosine learning rate~\cite{loshchilov2016sgdr} schedule, which gradually decays the learning rate from its initial value to 1e-8.
We report the average accuracy score on the test set within three runs.

\input{table/ablation_prompt_sample_strategy}
\input{figures/ablation_components}

\subsection{Compared to Stat-of-the-Art Results}\label{sec:compare}

\noindent\textbf{SPT achieves excellent results under SSL backbones.}
Here we compare SPT with full fine-tuning, VPT~\cite{jia2022visual}, and GateVPT~\cite{yoo2023improving}, over two different self-supervised pre-training approaches: MAE~\cite{he2022masked} and MoCo-v3~\cite{chen2021mocov3}. GateVPT~\cite{yoo2023improving} is a recent adaptation method specifically designed for self-supervised pre-training models. 
Tab.~\ref{tab:sota_ssl} reports the comparisons. 
For the first time, we showcase that a \textit{prompt tuning} method can achieve such highly accurate results and can even compete with full fine-tuning. Under MAE pre-training, our method improves average accuracy by up to $10\%\sim 30\%$ relative to VPT, and outperforms full fine-tuning in 19 out of 24 cases while using less than 0.4\% of learnable parameters. Moreover, on those tasks with relatively poor performance, SPT is the only one that is even on par with full parameter tuning (e.g., 80.13\% \textit{vs.} 80.55\% on CUB-200-2011 and 76.28\% \textit{vs.} 77.87\% on NABirds). SPT has also achieved large gains against VPT under MoCo-v3 pre-training, e.g., +2.88\% on FGVC \textit{Mean Acc} and +15.98\% on VTAB \textit{Structured} group.

In addition, we note that MoCo-v3 pre-training presents better adaptability to downstream tasks than MAE pre-training. Improving the adaptability of MAE pre-training is an interesting future topic.

\noindent\textbf{SPT is Comparable to PEFT Methods.}
Contrary to the scenario in self-supervised pre-training, these parameter-efficient fine-tuning techniques exhibit superior results compared to full fine-tuning in supervised pre-training. The current predominant approaches include optimizing the foundational backbones, such as SSF~\cite{lian2022scaling} and SNF~\cite{wang2023adapting}. We compare our SPT with previous methods in Tab.~\ref{tab:appendix_FGVC} with backbone ViT-B, which is supervised pre-trained on ImageNet-21K. In short, our method, SPT-deep, achieves the best mean accuracy on FGVC benchmarks (91.40\%). For more related experiments, please refer to Appendix Table.~\ref{tab:sota_sup}.




\subsection{Ablation Study and Analysis}\label{sec:ablation}
We perform ablation experiments on the FGVC and/or VTAB-1K dataset and compare them to the baseline that VPT~\cite{jia2022visual} with random initialization.
We set the prompt length to 100 and 20 for SPT-Shallow and SPT-Deep variants, respectively, with pre-trained ViT-B as the backbone. 

\noindent\textbf{The Impact of Different Sampling Strategies.}
We analyze the prompt sampling procedures when constructing prompts, as shown in Tab.~\ref{tab:ablation_prompt_sample}. We assess the performance of SPT-deep using two distinct pre-training methods and two separate recognition tasks to explore the impact of the prompt construction. In summary, our proposed approaches significantly outperform the existing strategy (random initialization in VPT).

Tab.~\ref{tab:ablation_prompt_sample_ptuning} provides a comparison between the k-means/mean pooling/max pooling/random sample strategies against the random initialization baseline. The setting is commonly used in prompt tuning that the prompts are optimized throughout the fine-tuning process. The values in the table demonstrate that not only the token prototypes can achieve good results, but also the proposed streamlined token construction approaches (Mean/Max pooling and Random Sample, more time cost can be found in Tab.~\ref{tab:time-cost}) work well.

We also study an interesting setting in Tab.~\ref{tab:ablation_prompt_sample_linear}, where the prompts are frozen after initialization. This means that only the task head is learnable during fine-tuning time (similar to linear probing). We found that our SPT-Deep can still achieve significant gains.

\noindent\textbf{SPT is Robust to Prompt Length.}
Fig.~\ref{fig:ablation_componments}~(c) investigates the influence of prompt length on accuracy. In this analysis, we vary the number of prompts inserted into each block for both SPT-Deep and VPT-Deep. Two key observations emerge from our study. Firstly, as the length of prompts increases, VPT exhibits fluctuating behavior. In contrast, our proposed method, SPT, demonstrates robustness to changes in prompt length and even yields slight performance gains as the prompt length increases. A notable gap of approximately 15\% is observed between the optimal prompt length (i.e., $N_p=80$) and the worst case scenario ($N_p=100$) for VPT. Secondly, under the optimal prompt length, SPT achieves a significant improvement of +10\% with $48\times$ fewer prompt parameters (ViT-B has 12 blocks) compared to VPT. We deduce that the superior performance of SPT can be attributed to the utilization of more appropriate prompts.

\noindent\textbf{\textit{"Right Tool for the Right Job"}.}
In Fig.\ref{fig:ablation_componments}(b), we investigate the utilization of cross-task data for prompt construction. Specifically, we designate one of the five FGVC benchmarks as the downstream task, while employing data from the other tasks to form prompts. The results within the same column in Fig.\ref{fig:ablation_componments}~(b) are normalized for clarity. The ablation analysis reveals that prompts are more effectively constructed using data from the downstream task itself. Constructing prompts with data from other tasks results in suboptimal performance. We hypothesize that the prompts and image embedding features may require approximate distribution to adapt downstream task data features. This observation is intriguing and lays the groundwork for future research aimed at quantifying the transferability of models across diverse datasets.

\noindent\textbf{SPT Scales Well with Large Capacity.}
In Fig.\ref{fig:ablation_componments}(d), we conduct a comparison between SPT and VPT by scaling up the pre-trained backbone size. Overall, our method, SPT, exhibits superior scaling behavior. In contrast, VPT encounters degradation issues, as evident in the comparison between VPT with ViT-H and ViT-L. Notably, the significant advantages of SPT over VPT persist as the model scale increases. For example, SPT with ViT-H outperforms VPT with ViT-H by up to 15.5\%.

\input{figures/fraction_train_data}

\noindent\textbf{Insights for Downstream Tuning.}
In Tab.~\ref{tab:sota_ssl}, our method, along with other parameter-efficient fine-tuning methods, demonstrates superior results compared to full fine-tuning. However, we emphasize that the task data used is relatively small, with only 1,000 training images in VTAB-1K, making full fine-tuning more susceptible to heavy overfitting.
Here, we employ ImageNet-1K (approximately 1.28 million training images) as the downstream task and vary the number of available training images to investigate the choice of fine-tuning methods. The results are presented in Fig.\ref{fig:fraction}. Similar observations were made on iNaturalist, a long-tail recognition task, as illustrated in Appendix (Fig.\ref{fig:Appendix_fraction}). We can easily obtain some insights: \textbf{(I):} With MAE pre-training, SPT and VPT outperform full fine-tuning significantly in scenarios with less task data (e.g., less than 10\%). However, the situation reverses with more task data. We infer that, with sufficient task data, SPT, VPT, and other parameter-efficient fine-tuning methods cannot adequately capture data information with only a few learnable parameters (usually less than 1\% of Full). A similar observation holds for supervised pre-training. \textbf{(II):} Supervised pre-training is a preferable choice when the available task data is small, despite the difference in pre-training data scale between MAE and supervised. Supervised pre-training can achieve satisfactory results with only minimal task data, e.g., 1\%. We infer that the supervised pre-trained backbone already possesses recognition capabilities (knowledge), thereby reducing the reliance on downstream task data.

\section{Discussion and Conclusion}
\textbf{Limitations.}
Our work has some visible limitations, e.g., we do not provide rigorous theoretical guarantee for the proposed initialization strategy and look forward to future research endeavors for this work.


\textbf{Conclusion.}
Our work proposes a simple and novel method (termed \textbf{SPT}) to improve visual prompt tuning. Specifically, we observe that the mutual information between prompt tokens and image patch tokens gradually increases during the prompt tuning process of VPT, and propose the idea of constructing prompts based on inferred token prototypes. This idea can make the prompt tokens and patch tokens share larger mutual information at the beginning of training, which benefits the rapid convergence of the model and achieve higher accuracy. Our method delivers substantial performance improvements, especially for self-supervised pre-trained models. 
For instance, after MAE pre-training, our method improves accuracy by up to 10\%$\sim$30\% compared to VPT, and outperforms Full fine-tuning 19 out of 24 cases while using less than 0.4\% of learnable parameters. Significant performance improvements are also achieved for MoCo-V3 and supervised pre-training. We conduct sufficient ablation experiments and analyses. The results show that our method, SPT, exhibits better than VPT (a strong baseline), e.g., SPT is robust to prompt length and achieves slight gains with increasing prompt length; Also, SPT scales well with model capacity. These characteristics still work well with supervised pre-trained backbone.

\section*{Acknowledgements}
The authors would like to thank the anonymous reviewers for their valuable comments and suggestions.
This work was supported by the National Key R\&D Program of China under Grant No. (2022YFF0608000, 2022ZD0119004), by the Fundamental Research Funds for the Central Universities, in part by the National Natural Science Foundation of China under Grant No. (72188101, 62106235, 62376206, U21B2048, 62322605, 62293543).

\section*{Impact Statement}
This paper presents work whose goal is to advance the field of Machine Learning. There are many potential societal consequences of our work, none which we feel must be specifically highlighted here.

\bibliography{example_paper}
\bibliographystyle{icml2024}

\newpage
\appendix
\onecolumn
\setcounter{table}{0}
\setcounter{figure}{0}
\renewcommand{\thetable}{S\arabic{table}}
\renewcommand{\thefigure}{S\arabic{figure}}

\section{Detailed Descriptions for the Evaluation Datasets}
We follow the practice of VPT~\cite{jia2022visual} to perform the split of train/val/test. 
Tab.~\ref{tab:appendix_data_info} summarizes the details of the evaluated datasets used in the paper.

\input{table/appendix_data_info}
\vspace{-2mm}
\input{figures/Appendix_ablation_components}
\vspace{-5mm}
\section{Additional Results}
\subsection{Ablation with Supervised Pre-training}
In Fig.\ref{fig:appendix_ablation_componments_sup}, we present the counterpart of the ablation study conducted with ViT-B using supervised pre-training, complementing the analysis in Fig.\ref{fig:ablation_componments}. The observations align with those obtained from MAE pre-training experiments. When compared to the baseline of VPT utilizing random initialization for prompts, SPT demonstrates superior behavior. Notably, SPT showcases robustness to prompt length variations and exhibits improved scaling behavior.

\input{figures/prompt_init_strategy}
\input{table/ablation_prompt_init}

\subsection{Maximal Benefits via Layer-Specific Correspondence}

In Tab.\ref{tab:ablation_prompt_init}, we present a comparison of prompt initialization strategies as illustrated in Fig.\ref{fig:prompt_init_strategy}.


SPT demonstrates significant superiority over the baseline that randomly initialize the parameters of prompts, particularly in the case of Masked Autoencoder (MAE) pre-training. It increases accuracy by up to 28.9\% and 11.8\% for shallow and deep variants, respectively. It's important to note that prompts need to be inserted into the input space of the transformer block corresponding to the depth of SPT. To assess this design, we selectively use the output features of a specific level, such as the first or last layer's features, to construct prompts. These prompts are then inserted into all levels of the block, as depicted in Fig.~\ref{fig:prompt_init_strategy} (b, c).

Tab.~\ref{tab:ablation_prompt_init} (b, c) reveals that while using features from a single level to construct prompts achieves significant gains over the baseline with random initialization, it still lags notably behind the performance of SPT. We infer that the primary reason for this is the inconsistent semantic granularity between prompts and image embedding features at the same transformer depth.




\newpage
\subsection{Additional  Results on Varying Training Data Size}

\input{figures/Appendix_fraction_train_data}
In Sec~\ref{sec:ablation}, we employ ImageNet-1K, comprising approximately 1.28 million training images, as the downstream task. We systematically vary the quantity of available training images to investigate the optimal fine-tuning method selection, as illustrated in Fig.~\ref{fig:fraction}. Here, we extend our experiments to iNaturalist 2018, a dataset with approximately 437,000 training images distributed across 8,142 classes, showcasing a natural, long-tailed distribution. The results, presented in Fig.~\ref{fig:Appendix_fraction}, mirror those of Fig.~\ref{fig:fraction}. Specifically, Self-Prompt Tuning (SPT) outperforms full fine-tuning when training data is limited, and the situation reverses when ample training data is available.

\subsection{Representation Similarity Analysis}

We employ the Centered Kernel Alignment (CKA~\cite{kornblith2019similarity}) method to measure the representation similarity between prompts and image embeddings across various pre-training objectives. Our findings reveal that SPT consistently demonstrates a higher degree of similarity between prompts and image embeddings within the same layer, irrespective of whether the model is supervised or self-supervised pre-trained. Contrastingly, during self-supervised pre-training, VPT exhibits a relatively disordered phenomenon. Specifically, prompts from a certain layer in VPT may exhibit greater similarity to the image embeddings of other layers, rather than those of the same layer. This observed behavior could potentially be a primary factor contributing to the suboptimal performance of VPT under self-supervised pre-training.

\input{figures/Appendix_CKA}

\newpage
\section{Per-task Results on VTAB-1K and FGVC}
Tab.~\ref{tab:appendix_vtab} and \ref{tab:sota_sup} present pre-task results for 24 classification tasks evaluated in the main paper.
\input{table/Appendix_VTAB}

\input{table/compare_sup}


\end{document}

%% file: figures/freamwork.tex
\begin{figure}[t]
\centering
    \includegraphics[width = 0.48\textwidth]{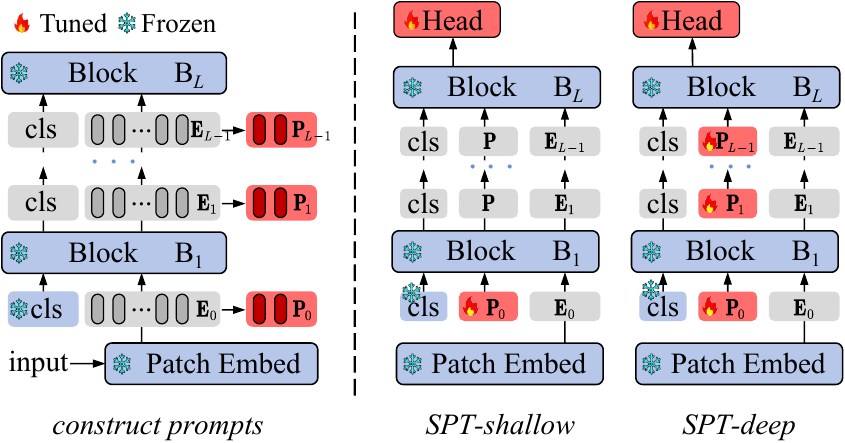}
\vspace{-7mm}
\caption{
\small
\textbf{Self-Prompt Tuning.}
\textbf{Left}: We input a batch of the training data from the downstream task into the pre-trained model to get the forward patch embeddings.
\textbf{Right}: We initialize prompts with sampled patch embeddings. Similar to VPT, we proposed  \textbf{SPT-Shallow} and \textbf{SPT-Deep} depending on the layers involved.
Only the prompts and task head parameters are learnable during adaptation on downstream tasks while the transformer encoder is frozen.
}
\vspace{-5mm}
\label{fig:freamwork}
\end{figure}

%% file: figures/nmi.tex
\begin{figure}
\begin{minipage}[h]{0.48\textwidth}
  \centering
  \centerline{
  \includegraphics[width = 0.54\textwidth, height=1.52cm]{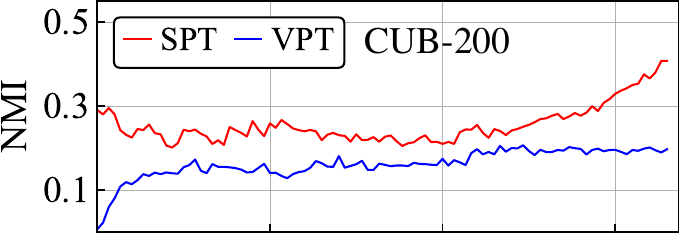}
  \hfill
  \includegraphics[width = 0.46\textwidth, height=1.52cm]{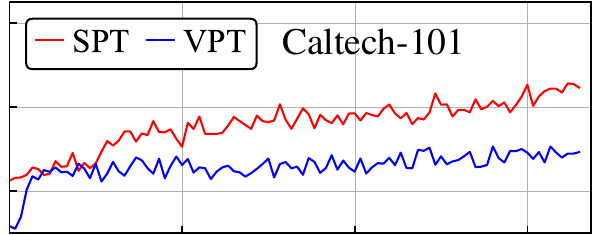}
  }
\end{minipage}

\begin{minipage}[h]{0.48\textwidth}
  \centering
  \centerline{
  \includegraphics[width = 0.54\textwidth, height=2.1cm]{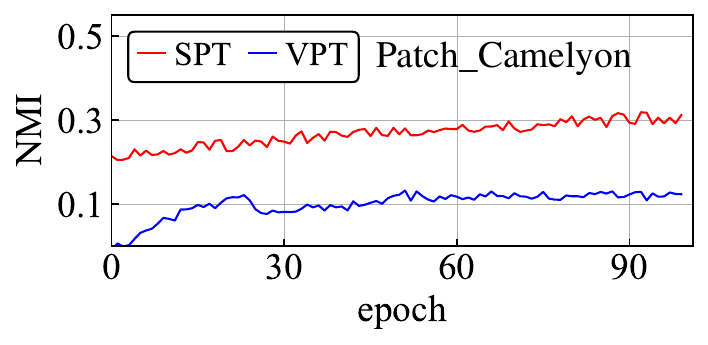}
  \hfill
  \includegraphics[width = 0.46\textwidth, height=2.1cm]{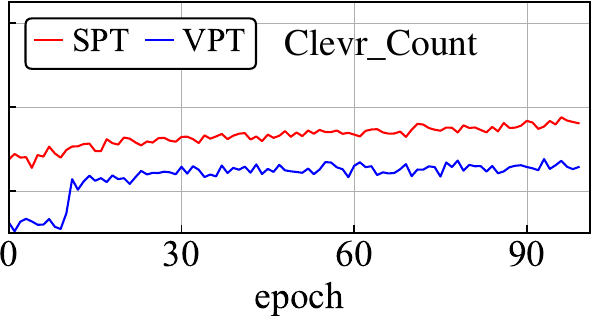}
  }
\end{minipage}
\vspace{-3mm}
\caption{VPT presents the behavior of the \textbf{N}ormalized \textbf{M}utual \textbf{I}nformation (NMI~\cite{estevez2009normalized}) between prompts and patch tokens gradually increases during fine-tuning time. 
SPT has large NMI at the beginning, which will facilitate rapid convergence and achieve more advanced results.}
\vspace{-4mm}
\label{fig:nmi}
\end{figure}

%% file: table/compare_ssl.tex
\begin{table*}[h]
\centering
\caption{\small 
\textbf{Comparison under self-supervised pre-trained backbones.}
We compare SPT and previous methods on 5 FGVC and 19 VTAB-1K benchmarks over two types of self-supervised pre-training methods: MAE~\cite{he2022masked} and MoCo-v3~\cite{chen2021mocov3}.
In short, our method achieves gains of 10\% to 30\% in average accuracy compared to VPT, and even outperforms full fine-tuning in 19 out of 24 cases while using less than 0.4\% of learnable parameters under MAE pre-training. Per-task results for VTAB-1K are presented in Appendix Table.~\ref{tab:appendix_vtab}.
}
\renewcommand{\arraystretch}{1.2}
\setlength\tabcolsep{2.0pt}
\resizebox{1.0\textwidth}{!}{%
    \footnotesize
    \begin{tabular}{l|cccccc|ccc}
        ~ & \multicolumn{6}{c}{FGVC} & \multicolumn{3}{|c}{VTAB-1K}\\
        Methods & Mean Acc & CUB-200-2011 & NABirds & Oxford Flowers & Stanford Dogs & Stanford Cars & \textit{Natural} (7) & \textit{Specialized} (4) & \textit{Structured} (8)\\
        \hline
        \toprule
        \multicolumn{10}{c}{\textit{ViT-B with MAE pretrained on ImageNet-1K}} \\
        Full              & 82.80 & \textbf{80.55} & \textbf{77.87} & 91.71 & 80.38 & 83.51 & 59.31 & 79.68 & 53.82\\
        VPT-Shallow       & 57.84 & 42.15 & 57.43 & 69.15 & 77.07 & 43.38 & 39.96 & 69.65 & 27.50\\
        VPT-Deep          & 72.02 & 68.33 & 65.22 & 80.05 & 78.83 & 67.67 & 36.02 & 60.61 & 26.57\\
        GateVPT           & 73.39 & 70.56 & 67.26 & 78.55 & 78.90 & 71.70 & 47.61 & 76.86 & 36.80\\
        SPT-Shallow (ours)& 73.95 & 71.15 & 61.87 & 89.47 & 80.01 & 67.23 & 62.53 & 80.90 & 53.46\\
        SPT-Deep (ours)   & \textbf{83.26} & 80.13 & 76.28 & \textbf{93.07} & \textbf{82.23} & \textbf{84.61} & \textbf{67.19}  & \textbf{83.15} & \textbf{59.23}\\
        \hline
        \multicolumn{10}{c}{\textit{ViT-B with MoCo-V3 pretrained on ImageNet-1K}} \\
        Full               & 84.25 & 81.75 & \textbf{78.14} & 94.52 & 81.19 & 85.67 & 71.95 & 84.72 & 51.98\\
        VPT-Shallow        & 79.26 & 79.05 & 72.92 & 90.47 & 81.97 & 71.91 & 67.34 & 82.26 & 37.55\\
        VPT-Deep           & 83.12 & 82.67 & 75.99 & 94.41 & 83.33 & 79.18 & 70.27 & 83.04 & 42.38\\
        GateVPT            & 83.00 & 82.86 & 76.02 & 93.71 & 83.37 & 79.02 & 74.84 & 83.38 & 49.10\\
        SPT-Shallow (ours) & 84.08 & 83.50 & 75.79 & 95.03 & 84.17 & 81.93 & 74.47 & 83.93 & 55.16\\
        SPT-Deep (ours)    & \textbf{86.00} & \textbf{84.47} & 77.63 & \textbf{96.10} & \textbf{85.84} & \textbf{85.98} & \textbf{76.20} & \textbf{84.95} & \textbf{58.36}\\
    \end{tabular}
}
\vspace{-5mm}
\label{tab:sota_ssl}
\end{table*}

%% file: table/Appendix_FGVC.tex
\begin{table*}[!htb]
\centering
\caption{
\textbf{Comparisons with PEFT approaches} with ViT-B pretrained on supervised ImageNet-21K as backbone.
We report the average performance (Mean Acc and Params) and per-task results.
“Input” and “Backbone” indicate the tuning parameter scope of each method.
SPT achieves competitive results with far fewer trainable parameters.
More results are presented in the Appendix (Table.~\ref{tab:appendix_vtab} and \ref{tab:sota_sup}).
}
\renewcommand{\arraystretch}{1.2}
\setlength\tabcolsep{4.8pt}
\resizebox{1.0\textwidth}{!}{%
    \footnotesize
    \begin{tabular}{l|cccc|ccccc}
        Methods & Input & Backbone & Params (M) & Mean Acc & CUB-200 & NABirds & Flowers & Dogs & Cars\\
        \hline
        \toprule
        Full & ~ & $\checkmark$ & 85.98 & 88.54 & 87.3 & 82.7 & 98.8 & 89.4 & 84.5\\
        Linear & ~ & ~          & 0.18  & 79.32 & 85.3 & 75.9 & 97.9 & 86.2 & 51.3\\
        \hline
        Bias~\cite{zhai2019large} & ~ & $\checkmark$ & 0.28  & 88.41 & 88.4 & 84.2 & 98.8 & \textbf{91.2} & 79.4\\
        Adapter\cite{houlsby2019parameter} & ~ & $\checkmark$ & 0.41 & 85.66 & 87.1 & 84.3 & 98.5 & 89.8 & 68.6\\
        MP~\cite{gao2023tuning}  & ~ & $\checkmark$  & 1.20 & 89.38 & 89.3 & 84.9 & 99.6 & 89.5 & 83.6\\
        SSF~\cite{lian2022scaling} & ~ & $\checkmark$  & 0.39 & 90.72 & 89.5 & 85.7 & 99.6 & 89.6 & 89.2\\
        SNF~\cite{wang2023adapting} & ~ & $\checkmark$  & 0.25 & 90.74 & 90.2 & 87.4 & 99.7 & 89.5 & 86.9\\
        \hline
        VPT-shallow~\cite{jia2022visual} & $\checkmark$ & ~ & 0.25 & 84.62 & 86.7 & 78.8 & 98.4 & 90.7 & 68.7\\
        VPT-Deep & $\checkmark$ & ~    & 0.85 & 89.11 & 88.5 & 84.2 & 99.0 & 90.2 & 83.6\\
        E$^2$VPT~\cite{cheng2023e2vpt} & $\checkmark$ & ~    & 0.56 & 89.22 & 89.1 & 84.6 & 99.1 & 90.5 & 82.8\\
        \hline
        SPT-Shallow (ours) & $\checkmark$ & ~    & 0.25 & 90.10  & 90.2 & 85.1 & 99.5 & 89.3 & 86.4\\
        SPT-Deep (ours) & $\checkmark$ & ~    & 0.36    & \textbf{91.40}  & \textbf{90.6} & \textbf{87.6} 
        & \textbf{99.8} & 89.8 & \textbf{89.2}\\
    \end{tabular}
}
\vspace{-1mm}
\label{tab:appendix_FGVC}
\end{table*}

%% file: table/ablation_prompt_sample_strategy.tex
\begin{table*}[t]
\centering
\caption{
\textbf{Ablation on prompt sampling strategies} with ViT-B as the backbone. 
We perform evaluation under two different pre-training strategies and two different types of downstream tasks: fine-grained (CUB-200, one of the FGVC benchmarks) and general natural images (Caltech-101, one of the VTAB-1K benchmarks), to explore the impact of sampling strategies.
Experimental results (\%) show that all prompt sampling strategies are significantly superior to the VPT baseline with random initialization~\cite{jia2022visual}.
}
\vspace{2mm}
    \begin{subtable}[h]{0.49\textwidth}
        \renewcommand{\arraystretch}{1.0}
        \setlength\tabcolsep{1.5pt}
        \fontsize{8.3}{\baselineskip}\selectfont
        \centering
        \caption{\textbf{The parameters of prompts are frozen} after initialization. Only the parameters of the task head are learnable during adaptation on downstream tasks.}
        \begin{tabular}{l|cc|cc}
            ~ & \multicolumn{2}{c}{IN-21K, sup} & \multicolumn{2}{|c}{IN-1K, MAE}\\
            prompt strategy & CUB-200 & Caltech-101 & CUB-200 & Caltech-101\\
            \hline
            \toprule
            Random init (VPT) & 81.6 & 81.9 & 25.3 & 73.5\\
            \hline
            K-means & 85.1 (\textcolor{darkgreen}{+3.5}) & 87.8 (\textcolor{darkgreen}{+5.9}) & 39.2 (\textcolor{darkgreen}{+13.8}) & 84.1 (\textcolor{darkgreen}{+10.6})\\
            Mean pooling & 83.5 (\textcolor{darkgreen}{+1.9}) & 86.3 (\textcolor{darkgreen}{+4.4}) 
            & 37.7 (\textcolor{darkgreen}{+12.4}) & 81.5 (\textcolor{darkgreen}{+8.0}) \\
            Max pooling & 84.0 (\textcolor{darkgreen}{+2.4}) & 87.1 (\textcolor{darkgreen}{+5.2})   & 37.0 (\textcolor{darkgreen}{+11.7}) & 81.6 (\textcolor{darkgreen}{+8.1})\\
            Random sample & 84.2 (\textcolor{darkgreen}{+2.6}) & 86.9 (\textcolor{darkgreen}{+5.0}) & 37.8 (\textcolor{darkgreen}{+12.5}) & 82.2 (\textcolor{darkgreen}{+8.7})\\
        \end{tabular}
        \label{tab:ablation_prompt_sample_linear}
    \end{subtable}
    \hfill
    \begin{subtable}[h]{0.49\textwidth}
        \renewcommand{\arraystretch}{1.0}
        \setlength\tabcolsep{1.5pt}
        \fontsize{8.3}{\baselineskip}\selectfont
        \centering
        \caption{During adaptation on downstream tasks, the parameters of prompts and task heads are learnable. This is a commonly used manner for visual prompt tuning.}
        \begin{tabular}{l|cc|cc}
            ~ & \multicolumn{2}{c}{IN-21K, sup} & \multicolumn{2}{|c}{IN-1K, MAE}\\
            prompt strategy & CUB-200 & Caltech-101 & CUB-200 & Caltech-101\\
            \hline
            \toprule
            Random init (VPT)& 88.5 & 90.8 & 68.3 & 80.2\\
            \hline
            K-means & 91.2 (\textcolor{darkgreen}{+2.7}) & 92.7 (\textcolor{darkgreen}{+1.9}) & 79.9 (\textcolor{darkgreen}{+11.6}) & 89.1 (\textcolor{darkgreen}{+8.9})\\
            Mean pooling & 90.4 (\textcolor{darkgreen}{+1.9}) & 92.5 (\textcolor{darkgreen}{+1.7}) & 79.8 (\textcolor{darkgreen}{+11.5}) & 88.6 (\textcolor{darkgreen}{+8.4})\\
            Max pooling & 90.6 (\textcolor{darkgreen}{+2.1}) & 92.1 (\textcolor{darkgreen}{+1.3}) & 79.6 (\textcolor{darkgreen}{+11.3}) & 88.7 (\textcolor{darkgreen}{+8.5})\\
            Random sample & 90.6 (\textcolor{darkgreen}{+2.1}) & 92.6 (\textcolor{darkgreen}{+1.8}) & 80.1 (\textcolor{darkgreen}{+11.8}) & 88.9 (\textcolor{darkgreen}{+8.7})\\
        \end{tabular}
        \label{tab:ablation_prompt_sample_ptuning}
    \end{subtable}
\vspace{-2mm}
\label{tab:ablation_prompt_sample}
\end{table*}

%% file: figures/ablation_components.tex
\begin{figure*}[!htb]
\centering
    \subcaptionbox{k-means, acc \textit{vs.} time}
    {\includegraphics[width = 0.24\textwidth, height=3.4cm]{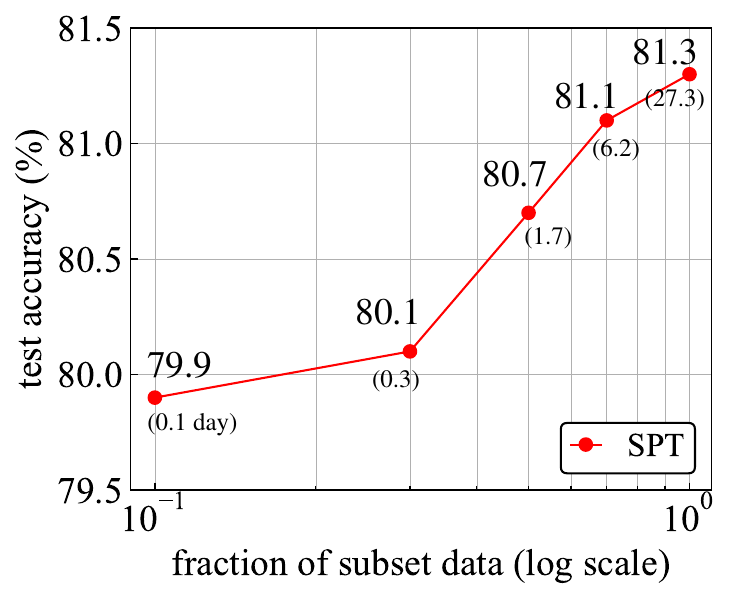}}
    \hfill
    \subcaptionbox{construct prompts across task}
    {\includegraphics[width = 0.24\textwidth, height=3.4cm]{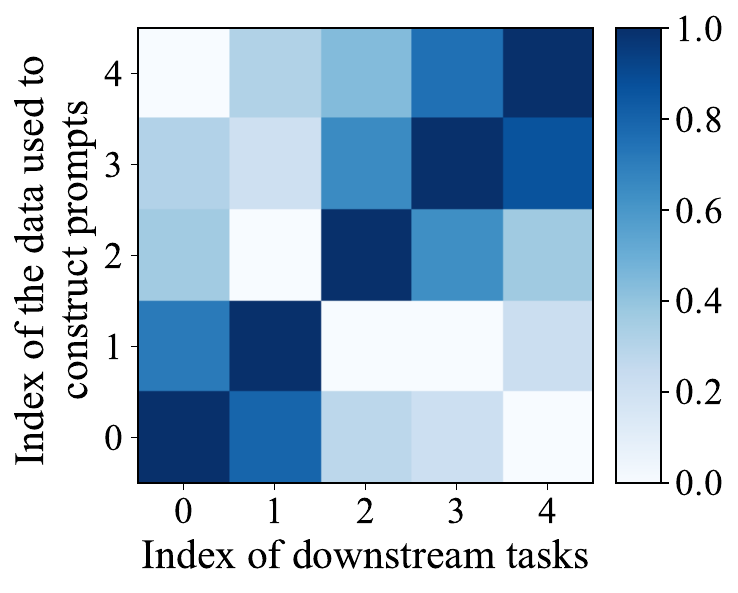}}
    \hfill
    \subcaptionbox{prompt length of each block}
    {\includegraphics[width = 0.24\textwidth, height=3.4cm]{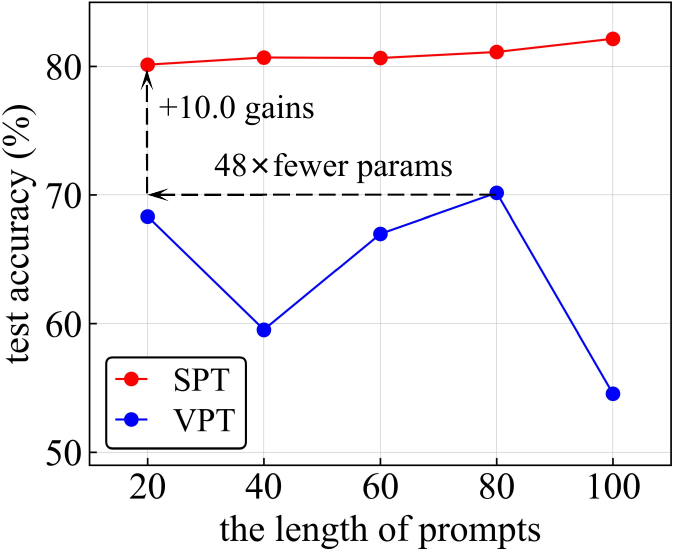}}
    \hfill
    \subcaptionbox{scale up model size}
    {\includegraphics[width = 0.24\textwidth, height=3.4cm]{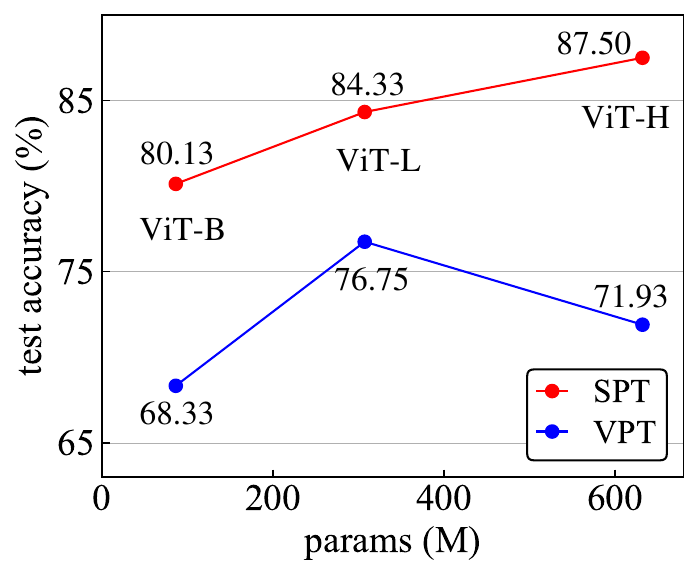}}
\caption{
\small
\textbf{Ablation studies on several basic components} using Masked Autoencoder (MAE) pre-trained backbone and SPT-deep evaluated on CUB-200-2011.
\textbf{(a)} The k-means cluster strategy achieves further improvement with more data used to construct prompts. However, it introduces significant time costs during construct prompts. The wall-clock time is displayed in ($\cdot$).
\textbf{(b)} Prompts should be constructed using data from downstream tasks. 
\textbf{(c)} 
SPT is robust to prompt length changes and achieves slight gains with increasing prompt length.
\textbf{(d)} SPT presents better scaling behavior than VPT with scaling up model size.
These observations under supervised pre-trained backbone are similar (see the Appendix).
}
\label{fig:ablation_componments}
\end{figure*}

%% file: figures/fraction_train_data.tex
\begin{figure}[t]
\centering
    \subcaptionbox{IN-1K, MAE}
    {\includegraphics[width = 0.238\textwidth]{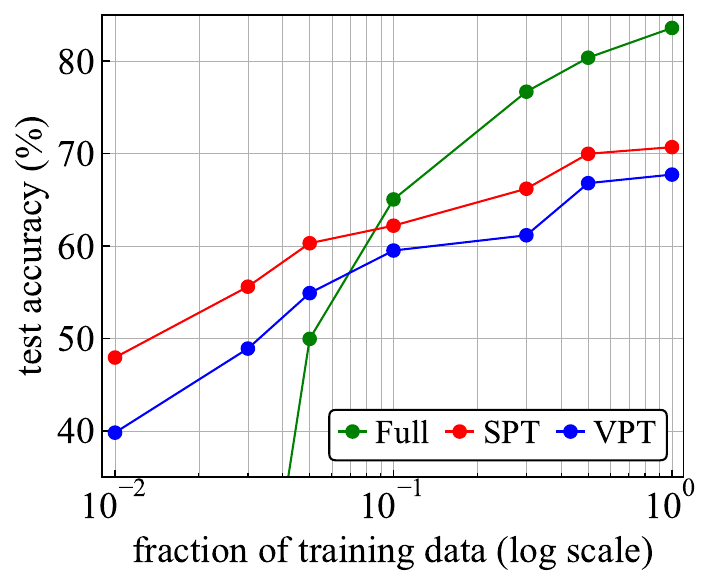}}
    \hfill
    \subcaptionbox{IN-21K, sup}
    {\includegraphics[width = 0.238\textwidth]{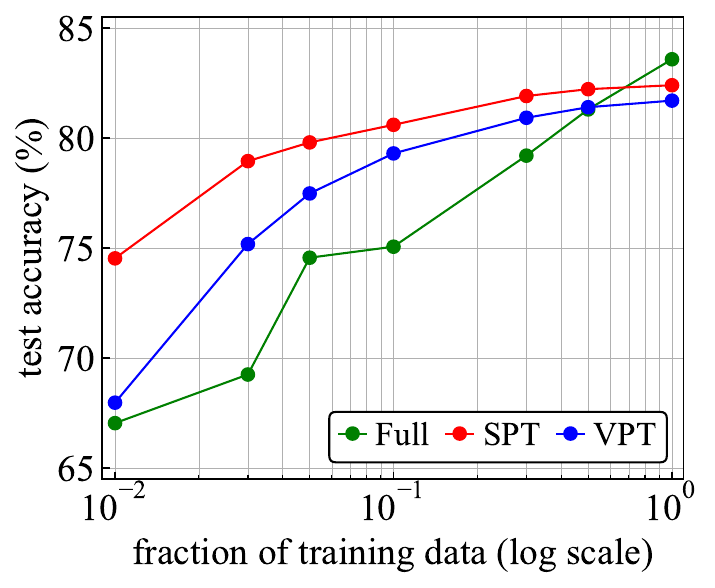}}
\caption{
\small
\textbf{The impact of varying tuning data sizes with MAE and supervised pretraining.
}}
\vspace{-4mm}
\label{fig:fraction}
\end{figure}

%% file: table/Appendix_data_info.tex
\begin{table}[h]
\vspace{-3mm}
\centering
\caption{\small 
\textbf{Specifications of the downstream task datasets evaluated.} We randomly sampled the \texttt{train} and \texttt{val} sets since there are no public splits available.}
\vspace{2mm}
\renewcommand{\arraystretch}{1.2}
\setlength\tabcolsep{5.0pt}
    \footnotesize
    \begin{tabular}{l|lllll}
        Datasets &  Description & \# Classes & Train & Val & Test\\
        \toprule
        \multicolumn{6}{c}{\textit{Fine-grained visual recognition tasks (FGVC)}} \\
        CUB-200-2011~\cite{wah2011caltech} & Fine-grained bird species recognition & 200 & 5,394$^*$ 
        & 600$^*$ & 5,794\\
        NABirds~\cite{van2015building} & Fine-grained bird species recognition & 555 & 21,536$^*$ & 2,393$^*$ & 24,633\\
        Oxford Flowers~\cite{nilsback2008automated} & Fine-grained flower species recognition & 102 & 1,020 & 1,020 & 6,149\\
        Stanford Dogs~\cite{khosla2011novel} & Fine-grained dog species recognition & 120 & 10,800$^*$
        & 1,200$^*$ & 8,580\\
        Stanford Cars~\cite{gebru2017fine} &  Fine-grained car classification & 196 & 7,329$^*$ & 815$^*$
        & 8,041\\
        \hline
        \multicolumn{6}{c}{\textit{Visual Task Adaptation Benchmark (VTAB-1k)}} \\
        Caltech101~\cite{fei2006one} & \multirow{7}{*}{Natural (7)} & 102 & \multirow{7}{*}{800 / 1000} & \multirow{7}{*}{200} & 6,084\\
        CIFAR-100~\cite{krizhevsky2009learning} & ~ & 100 & ~ & ~ & 10,000\\
        DTD~\cite{cimpoi2014describing}       & ~ & 47  & ~ & ~ & 1,880\\
        Flowers102~\cite{nilsback2008automated} & ~ & 102 & ~ & ~ & 6,149\\
        Pets~\cite{parkhi2012cats}      & ~ & 37  & ~ & ~ & 3,669\\
        SVHN~\cite{netzer2011reading} & ~ & 10  & ~ & ~ & 26,032\\
        Sun397~\cite{xiao2010sun}    & ~ & 397 & ~ & ~ & 21,750\\
        \hline
        Patch Camelyon~\cite{veeling2018rotation} & \multirow{4}{*}{Specialized (4)} & 2 & \multirow{4}{*}{800 / 1000} & \multirow{4}{*}{200} & 32,768\\
        EuroSAT~\cite{helber2019eurosat}   & ~ & 10 & ~ & ~ & 5,400\\
        Resisc45~\cite{cheng2017remote}  & ~ & 45 & ~ & ~ & 6,300\\
        Retinopathy~\cite{graham2015kaggle} & ~ & 5 & ~ & ~ & 42,670\\
        \hline
        Clevr/count~\cite{johnson2017clevr} & \multirow{8}{*}{Structured (8)} & 8 & \multirow{8}{*}{800 / 1000} & \multirow{8}{*}{200} & 15,000\\
        Clevr/distance~\cite{johnson2017clevr} & ~ & 6 & ~ & ~ & 15,000\\
        DMLab~\cite{beattie2016deepmind} & ~ & 6 & ~ & ~ & 22,735\\
        KITTI/distance~\cite{geiger2013vision} & ~ & 4 & ~ & ~ & 711\\
        dSprites/location~\cite{matthey2017dsprites} & ~ & 16 & ~ & ~ & 73,728\\
        dSprites/orientation~\cite{matthey2017dsprites} & ~ & 16 & ~ & ~ & 73,728\\
        SmallNORB/azimuth~\cite{lecun2004learning} & ~ & 18 & ~ & ~ & 12,150\\
        SmallNORB/elevation~\cite{lecun2004learning} & ~ & 9 & ~ & ~ & 12,150\\     
    \end{tabular}
\vspace{-2mm}
\label{tab:appendix_data_info}
\end{table}

%% file: figures/Appendix_ablation_components.tex
\begin{figure*}[!htb]
\centering
\begin{minipage}{0.9\textwidth}
    \centering
    \centerline{
    \subcaptionbox{construct prompts across task}
    {\includegraphics[width = 0.3\textwidth, height=3.8cm]{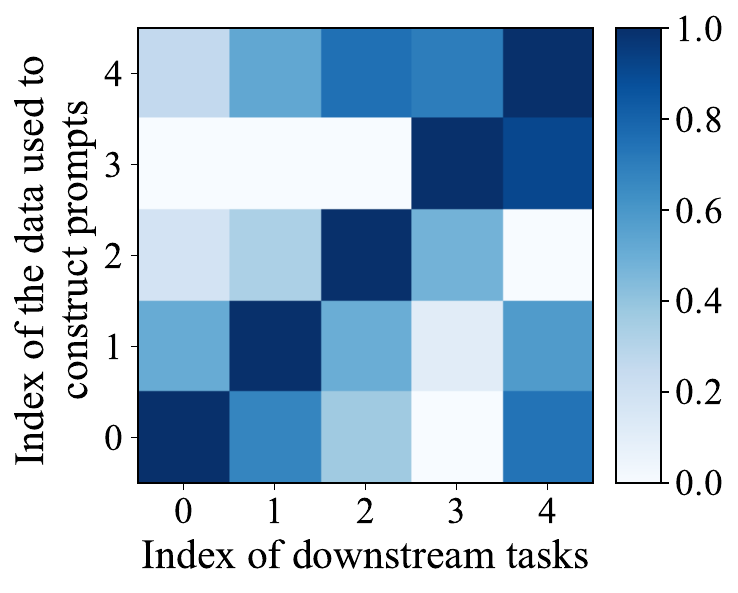}}
    \hfill
    \subcaptionbox{prompt length of each block}
    {\includegraphics[width = 0.3\textwidth, height=3.8cm]{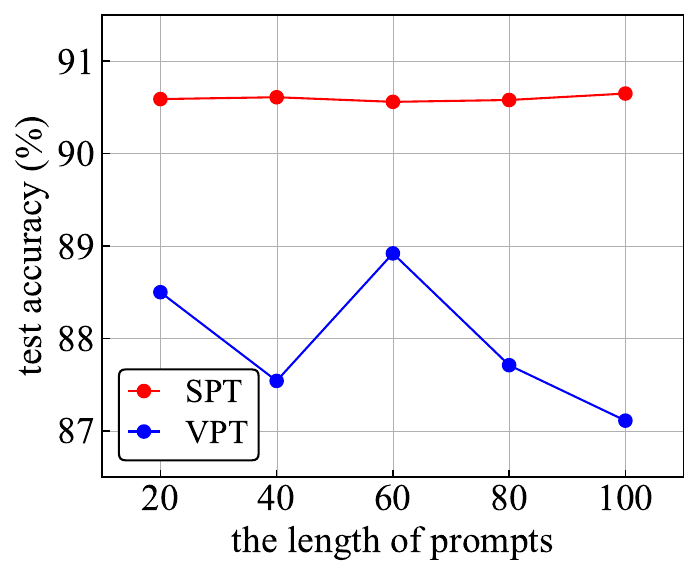}}
    \hfill
    \subcaptionbox{scale up model size}
    {\includegraphics[width = 0.3\textwidth, height=3.8cm]{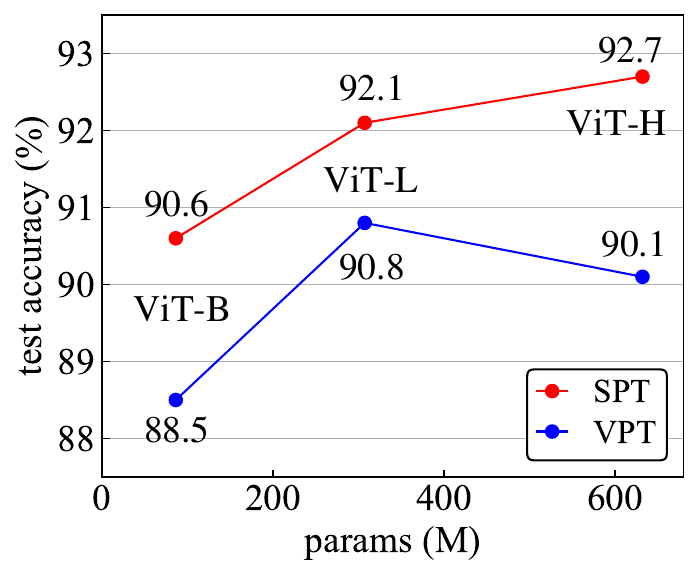}}
    }
\end{minipage}
\caption{
\small
\textbf{The IN-21K supervised} ViT-B counterpart of Fig.~\ref{fig:ablation_componments} on several major components.
}
\label{fig:appendix_ablation_componments_sup}
\end{figure*}

%% file: figures/prompt_init_strategy.tex
\begin{figure*}[t]
\centering
    \subcaptionbox{random initialization}
    {\includegraphics[width = 0.22\textwidth]{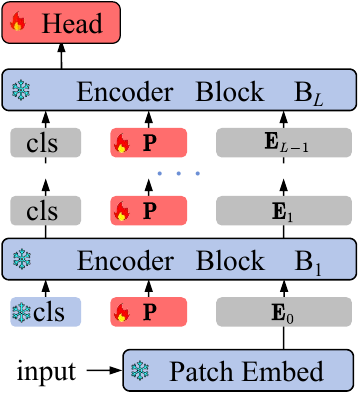}}
    \hfill
    \subcaptionbox{only use one-level's feature}
    {\includegraphics[width = 0.22\textwidth]{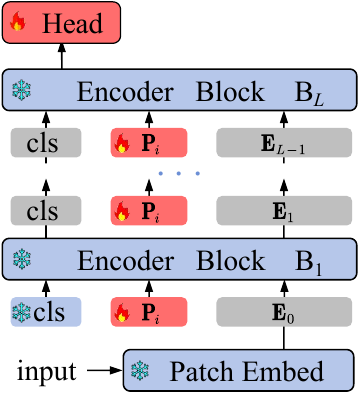}}
    \hfill
    \subcaptionbox{SPT-deep}
    {\includegraphics[width = 0.22\textwidth]{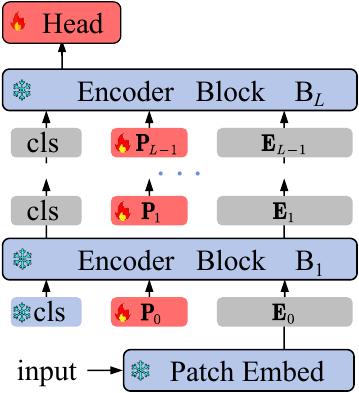}}
    \hfill
    \subcaptionbox{SPT-deep-frozen}
    {\includegraphics[width = 0.22\textwidth]{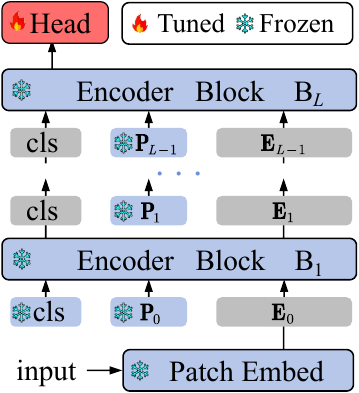}}
\caption{
\small
\textbf{Prompt initialization strategies} with pretrained backbones.
\textbf{(a)} Randomly initialize the parameters of prompts, which is adopted by VPT~\cite{jia2022visual} and its variants~\cite{yoo2023improving, cheng2023e2vpt, tu2023visual}.
\textbf{(b)} Only use the first (i.e., $\mathbf{P}_i=\mathbf{P}_0$) or the last ($\mathbf{P}_i=\mathbf{P}_L$) level's output features to initialize all levels of prompts.
\textbf{(c)} Use same-level image embedding features to construct prompts and insert the prompts into the input space of each block. We refer it as \textit{self-prompt} strategy.
\textbf{(d)} We found that after self-prompt initialization, even if the parameters of the prompt are frozen (which means that only the task head is learnable), our approach can still significantly improve the transfer performance (Table~\ref{tab:ablation_prompt_sample_linear}). In all four cases, the encoders are frozen.
}
\label{fig:prompt_init_strategy}
\end{figure*}

%% file: table/ablation_prompt_init.tex
\begin{table}[t]
\centering
\caption{\small 
\textbf{Ablation on prompt initialization strategies} with ViT-B as backbone evaluated on CUB-200-2011.
The backbone is initialized with supervised pre-training on ImageNet-21K (left) and MAE~\cite{he2022masked} pre-trained on ImageNet-1K without labels (right). Two variants, namely \textit{Shallow} and \textit{Deep}, are depicted in Fig.\ref{fig:freamwork}, with prompt lengths set to 100 and 20, respectively. Entries (a-c) correspond to Fig.\ref{fig:prompt_init_strategy} (a-c), with a comparison to the baseline represented by VPT with uniform initialization~\cite{jia2022visual}. The \textit{SPT} strategy outperforms the baseline, particularly when using the MAE pre-trained backbone. For the shallow variant, \textit{first P$_0$} and \textit{self-prompt} yield equivalent results.
}
\vspace{2mm}
\renewcommand{\arraystretch}{1.2}
\setlength\tabcolsep{9.0pt}
\resizebox{0.75\textwidth}{!}{%
    \small
    \begin{tabular}{ll|cc|cc}
        ~ & ~ & \multicolumn{2}{c}{IN-21K, sup} & \multicolumn{2}{|c}{IN-1K, MAE}\\
        ~ & prompt init. & Shallow & Deep & Shallow & Deep\\
        \hline
        \toprule
        (a) & random & 86.7 & 88.5 & 42.2 & 68.3 \\
        \hline
        \multirow{2}{*}{(b)} & first P$_0$ & 90.2 (\textcolor{darkgreen}{+3.5}) 
        & 89.2 (\textcolor{darkgreen}{+0.7}) & 71.1 (\textcolor{darkgreen}{+28.9}) 
        & 76.5 (\textcolor{darkgreen}{+8.2})\\
        ~ & last P$_L$ & 88.4 (\textcolor{darkgreen}{+1.7}) & 89.9 (\textcolor{darkgreen}{+1.4}) 
        & 64.6 (\textcolor{darkgreen}{+22.4}) & 76.8 (\textcolor{darkgreen}{+8.5})\\
        \hline
        (c) & SPT (self-prompt) & \textbf{90.2} (\textcolor{darkgreen}{+3.5}) 
        & \textbf{90.6} (\textcolor{darkgreen}{+2.1}) & \textbf{71.1} (\textcolor{darkgreen}{+28.9}) & \textbf{80.1} (\textcolor{darkgreen}{+11.8})\\
    \end{tabular}
}
\label{tab:ablation_prompt_init}
\end{table}

%% file: figures/Appendix_fraction_train_data.tex
\begin{wrapfigure}{r}{0.5\textwidth}
\vspace{-3mm}
\centering
    \subcaptionbox{MAE pre-training}
    {\includegraphics[width = 0.238\textwidth]{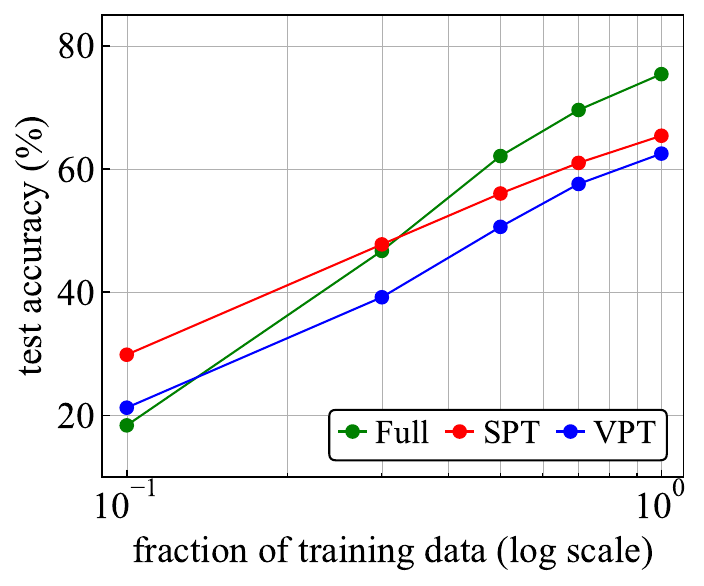}}
    \hfill
    \subcaptionbox{Sup pre-training}
    {\includegraphics[width = 0.238\textwidth]{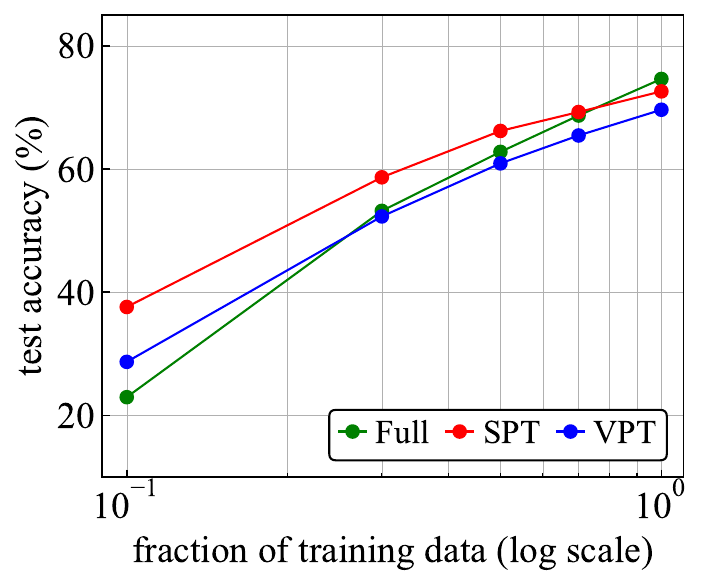}}
\caption{
\small
\textbf{Varying available training data on iNaturelist 2018} and across two different pre-traing objectives.
}
\label{fig:Appendix_fraction}
\end{wrapfigure}


%% file: figures/Appendix_CKA.tex
\begin{figure}[h]
\centering
\begin{minipage}[h]{0.9\textwidth}
  \centering
  \centerline{
  \includegraphics[width = 0.3\textwidth, height=4.7cm]{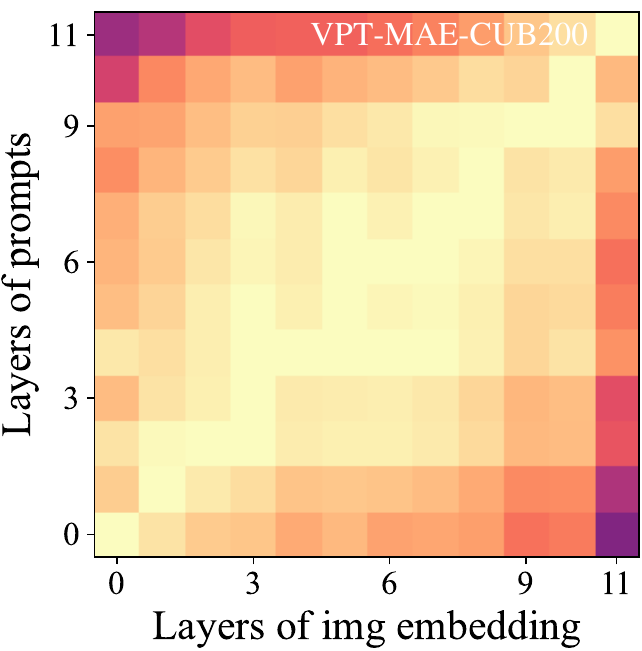}
  \hfill
  \includegraphics[width = 0.3\textwidth, height=4.7cm]{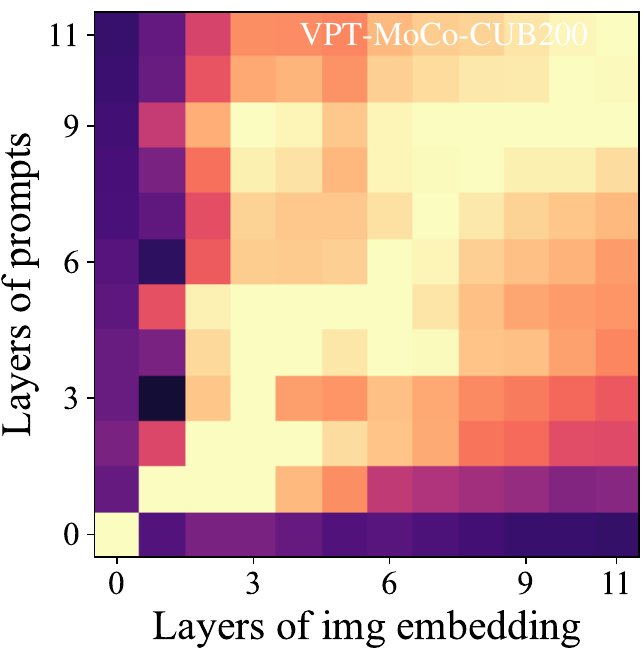}
  \hfill
  \includegraphics[width = 0.36\textwidth, height=4.7cm]{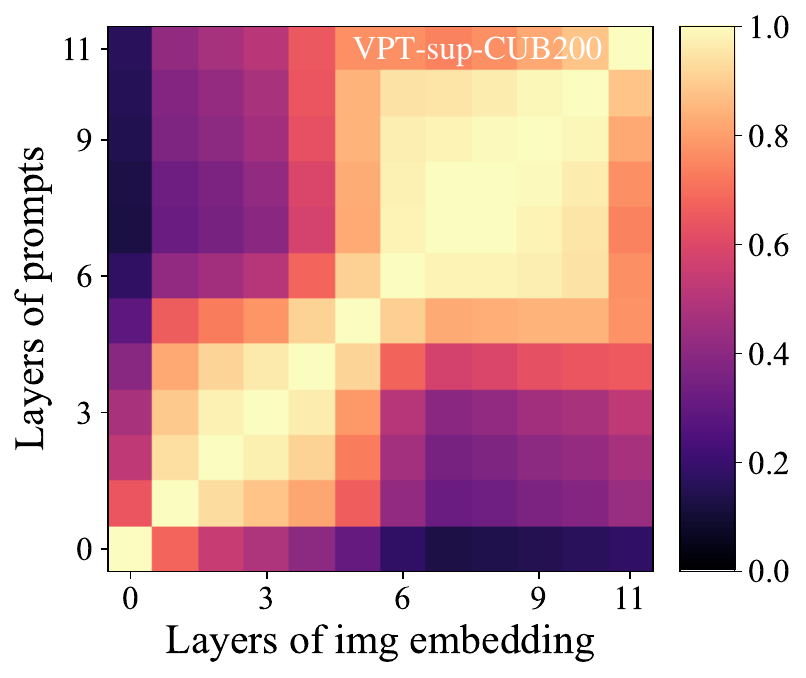}
  }
\end{minipage}

\begin{minipage}[h]{0.9\textwidth}
  \centering
  \centerline{
  \includegraphics[width = 0.3\textwidth, height=4.7cm]{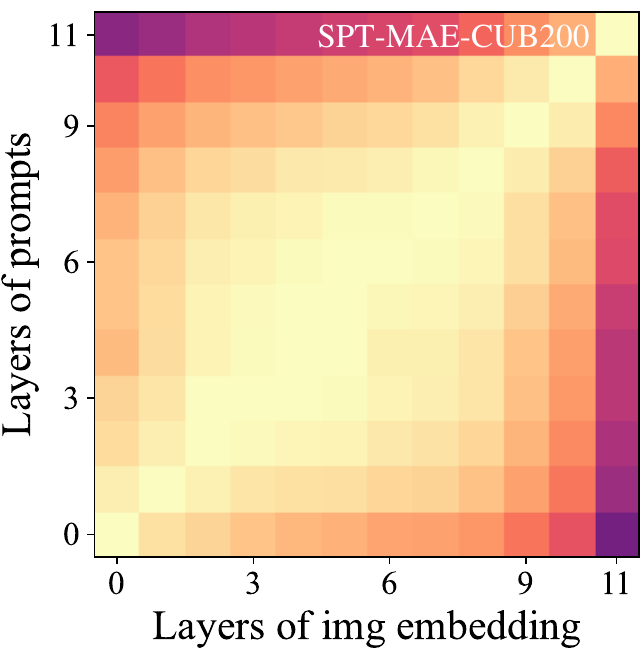}
  \hfill
  \includegraphics[width = 0.3\textwidth, height=4.7cm]{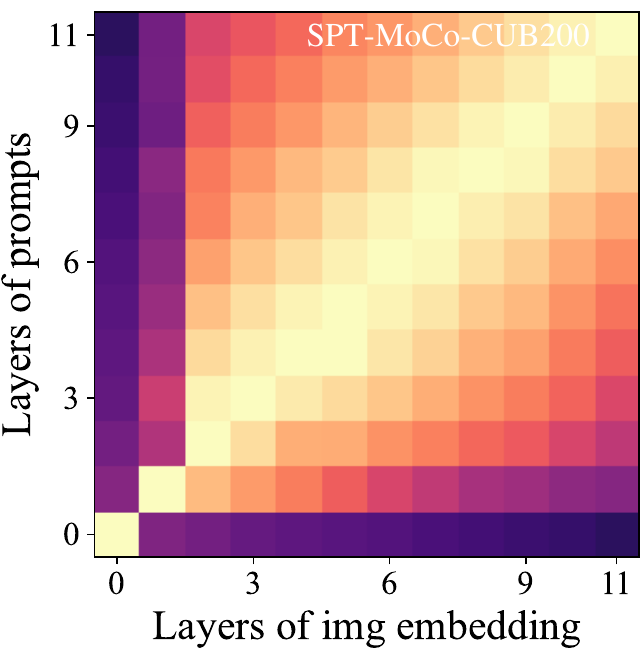}
  \hfill
  \includegraphics[width = 0.36\textwidth, height=4.7cm]{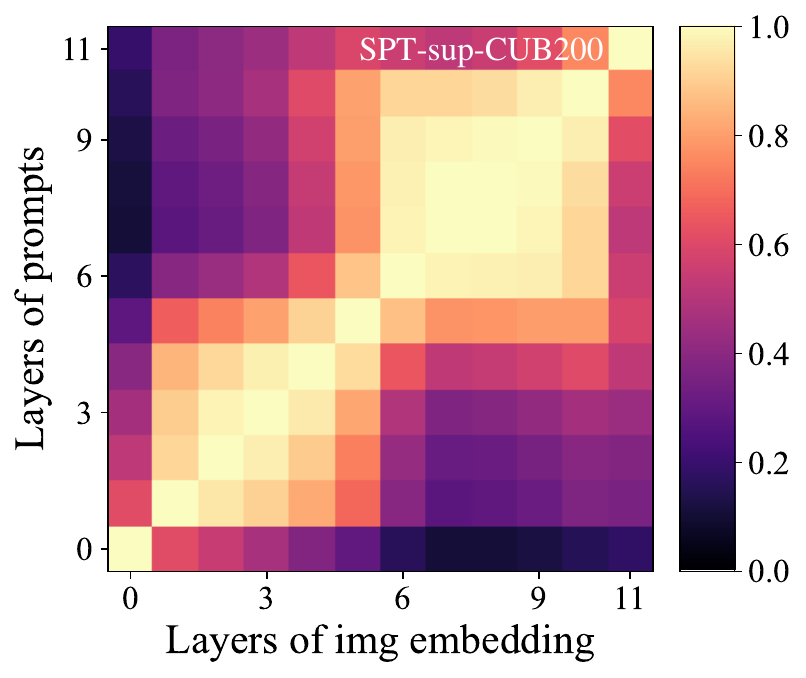}
  }
\end{minipage}
\caption{CKA similarities of prompts and patch tokens (i.e., image embedding features) for different layers and across different pre-training objectives.}
\label{fig:appendix_cka}
\end{figure}

%% file: table/Appendix_VTAB.tex
\begin{table*}[h]
\centering
\caption{\small 
Per-task fine-tuning results from Table.~\ref{tab:sota_ssl} for VTAB-1k benchmarks with pre-trained ViT-B/16 as backbone. The length of prompts is 100 and 20 for shallow and deep variants for all tasks respectively.
}
\vspace{2mm}
\renewcommand{\arraystretch}{1.2}
\setlength\tabcolsep{3.0pt}
\resizebox{1.0\textwidth}{!}{%
    \footnotesize
    \begin{tabular}{l|ccccccc|cccc|cccccccc}
        ~ & \multicolumn{7}{c}{\textit{Natural} (7)} & \multicolumn{4}{|c}{\textit{Specialized} (4)} & \multicolumn{8}{|c}{\textit{Structured} (8)}\\
        Methods & \rotatebox{90}{Caltech101} & \rotatebox{90}{CIFAR-100} & \rotatebox{90}{DTD} 
        & \rotatebox{90}{Flowers102} & \rotatebox{90}{Pets} & \rotatebox{90}{SVHN} & \rotatebox{90}{Sun397} 
        & \rotatebox{90}{Patch Camelyon} & \rotatebox{90}{EuroSAT} & \rotatebox{90}{Resisc45} 
        & \rotatebox{90}{Retinopathy} & \rotatebox{90}{Clevr/count} & \rotatebox{90}{Clevr/distance} 
        & \rotatebox{90}{DMLab} & \rotatebox{90}{KITTI/distance} & \rotatebox{90}{dSprites/loc} 
        & \rotatebox{90}{dSprites/ori} & \rotatebox{90}{SmallNORB/azi} 
        & \rotatebox{90}{SmallNORB/ele}\\
        \hline
        \toprule
        \multicolumn{19}{c}{\textit{ViT-B with MAE pretrained on ImageNet-1K}} \\
        Full               & 84.2 & 24.6 & 56.9 & 72.7 & 74.4 & \textbf{86.6} & 15.8 & 81.8 & \textbf{94.0} & 72.3 & 70.6 & 67.0 & 59.8 & 45.2 & 75.3 & 72.5 & 47.5 & \textbf{30.2} & 33.0\\
        SPT-Shallow (ours) & 87.6 & 29.5 & 61.5 & 77.4 & 80.8 & 77.2 & 23.7 & 84.4 & 93.0 & 70.8 & 75.4 & 73.3 & 55.5 & 44.0 & 73.2 & 70.6 & 48.0 & 27.4 & 35.7\\
        SPT-Deep (ours)    & \textbf{88.9} & \textbf{37.7} & \textbf{64.1} & \textbf{84.5} & \textbf{83.7} & 84.9 & \textbf{26.5} & \textbf{85.2} & 93.3 & \textbf{78.5} & \textbf{75.6} 
        & \textbf{76.8} & \textbf{63.1} & \textbf{47.9} & \textbf{76.7} & \textbf{82.3} & \textbf{49.3} & 29.6 & \textbf{48.1}\\
        \hline
        \multicolumn{19}{c}{\textit{ViT-B with MoCo-V3 pretrained on ImageNet-1K}} \\
        Full               & 91.0 & 57.6 & 64.6 & 91.6 & 79.9 & \textbf{89.8} & 29.1 & 85.1 & \textbf{96.4} & 83.1 & 74.2 & 55.2 & 56.9 & 44.6 & \textbf{77.9} & 63.8 & \textbf{49.0} & \textbf{31.5} & 36.9\\
        SPT-Shallow (ours) & 91.0 & 58.1 & 69.6 & 91.1 & 89.4 & 82.2 & 39.9 & 83.6 & 94.7 & 82.0 & 75.4 & 73.3 & 60.6 & 45.7 & 71.4 & 75.0 & 42.1 & 28.0 & 45.2\\
        SPT-Deep (ours)    & \textbf{92.2} & \textbf{62.3} & \textbf{70.1} & \textbf{93.2} & \textbf{89.4} & 84.3 & \textbf{41.9} & \textbf{85.6} & 95.1 & \textbf{83.3} & \textbf{75.8} & \textbf{76.1} & \textbf{62.5} & \textbf{49.0} & 77.3 & \textbf{77.7} & 45.6 & 29.5 & \textbf{49.2}\\
        \hline
        \multicolumn{19}{c}{\textit{ViT-B with supervised pretrained on ImageNet-21K}} \\
        Full               & 87.7 & 68.9 & 64.3 & 97.2 & 86.9 & 87.4 & 38.8 & 79.7 & 95.7 & 84.2 & 73.9 & 56.3 & 58.6 & 41.7 & 65.5 & 57.5 & 46.7 & 25.7 & 29.1\\
        SPT-Shallow (ours) & 91.2 & 78.9 & 71.2 & 99.4 & 90.7 & 86.4 & \textbf{52.3} & 82.5 
        & 94.9 & 82.6 & 74.6 & 68.6 & 61.2 & 48.0 & 68.8 & 72.6 & 41.9 & 24.4 & 37.4\\
        SPT-Deep (ours)    & \textbf{92.6} & \textbf{79.3} & \textbf{73.2} & \textbf{99.5} & \textbf{91.0} & \textbf{89.1} & 51.2 & \textbf{85.4} & \textbf{96.8} & \textbf{84.9} & \textbf{74.8} & \textbf{70.3} & \textbf{64.8} & \textbf{54.2} & \textbf{75.2} & \textbf{79.3} & \textbf{49.5} & \textbf{36.5} & \textbf{41.5}\\
    \end{tabular}
}
\vspace{-1mm}
\label{tab:appendix_vtab}
\end{table*}

%% file: table/compare_sup.tex
\begin{table*}[!htb]
\centering
\caption{\small 
Per-task fine-tuning results for FGVC and VTAB-1K benchmarks with pre-trained ViT-B/16 as backbone. The length of prompts is 100 and 20 for shallow and deep variants for all tasks, respectively.}
\vspace{2mm}
\renewcommand{\arraystretch}{1.2}
\setlength\tabcolsep{5.0pt}
\resizebox{1.0\textwidth}{!}{%
    \footnotesize
    \begin{tabular}{l|cc|cc|cccc}
        ~ & \multicolumn{2}{c}{Scope} & \multicolumn{2}{|c}{FGVC (5)} & \multicolumn{4}{|c}{VTAB-1K (19)}\\
        Methods & Input & Backbone & Params (M) & Mean Acc & Params (M) & \textit{Nature} (7) & \textit{Specialized} (4) & \textit{Structured} (8)\\
        \hline
        \toprule
        Full & ~ & $\checkmark$ & 85.98 & 88.54 & 85.84 & 75.88 & 83.36 & 47.64\\
        Linear & ~ & ~          &  0.18 & 79.32 &  0.04 & 68.93 & 77.16 & 26.84\\
        \hline
        Bias~\cite{zhai2019large} & ~ & $\checkmark$ & 0.28  & 88.41 & 0.14 & 73.30 & 78.25 & 44.09\\
        Adapter~\cite{houlsby2019parameter} & ~ & $\checkmark$ & 0.41 & 85.66 & 0.27 & 70.39 & 77.11 & 33.43\\
        LoRA~\cite{hu2021lora} & ~ & $\checkmark$      & -  & -  & 0.32 & 79.49 & 84.55 & \textbf{59.77}\\
        SSF~\cite{lian2022scaling} & ~ & $\checkmark$  & 0.39 & 90.72 & 0.24 & 81.57 & \textbf{86.55} & 58.96\\
        SNF~\cite{wang2023adapting} & ~ & $\checkmark$  & 0.25 & 90.74 & 0.29 & \textbf{83.78} & 86.13 & 59.61\\
        \hline
        VPT-shallow~\cite{jia2022visual} & $\checkmark$ & ~ & 0.25 & 84.62 & 0.11 & 76.81 & 79.68 & 46.98\\
        VPT-Deep~\cite{jia2022visual} & $\checkmark$ & ~    & 0.85 & 89.11 & 0.64 & 78.48 & 82.43 & 54.98\\
        E$^2$VPT~\cite{cheng2023e2vpt} & $\checkmark$ & ~    & 0.56 & 89.22 & 0.31 & 80.01 & 84.43 & 57.39\\
        \hline
        SPT-Shallow (ours) & $\checkmark$ & ~ & 0.25 & 90.10 & 0.11 & 81.44 & 83.65 & 52.86\\
        SPT-Deep (ours) & $\checkmark$ & ~    & 0.36 & \textbf{91.40} & 0.22 & 82.27 & 85.48 & 58.91\\
    \end{tabular}
}
\vspace{-5mm}
\label{tab:sota_sup}
\end{table*}